\documentclass[10pt,twocolumn,letterpaper]{article}

\usepackage{wacv}              

\usepackage{graphicx}
\usepackage{amsmath}
\usepackage{amssymb}
\usepackage{booktabs}
\usepackage{multirow}
\usepackage[normalem]{ulem}
\useunder{\uline}{\ul}{}

\usepackage[pagebackref,breaklinks,colorlinks]{hyperref}

\usepackage[capitalize]{cleveref}
\crefname{section}{Sec.}{Secs.}
\Crefname{section}{Section}{Sections}
\Crefname{table}{Table}{Tables}
\crefname{table}{Tab.}{Tabs.}

\def\wacvPaperID{1551} 
\def\confName{WACV}
\def\confYear{2025}

\begin{document}

\title{Volumetric Conditioning Module to Control Pretrained Diffusion Models \\ for 3D Medical Images}

\author{
    Suhyun Ahn \; Wonjung Park \; Jihoon Cho \; Seunghyuck Park \; Jinah Park \\
    KAIST \\
    {\tt\small \{ahn.ssu, fabiola, zinic, simonp6, jinahpark\}@kaist.ac.kr}
}

\twocolumn[{%
\renewcommand\twocolumn[1][]{#1}%
\maketitle
\begin{center}
    \centering
    \captionsetup{type=figure}
    \includegraphics[width=\textwidth]{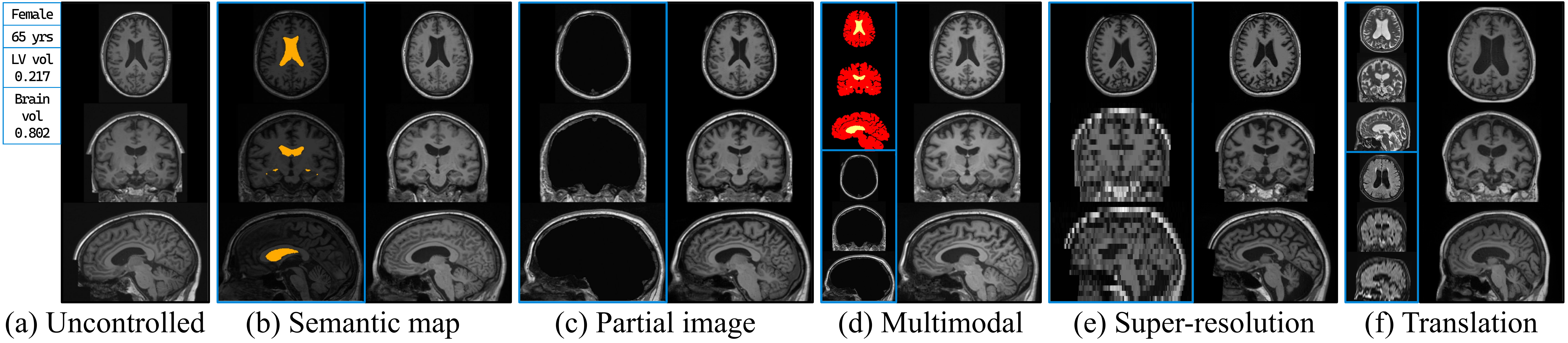}
    \captionof{figure}{Illustration of tasks using the proposed method, Volumetric Conditioning Module (VCM). Upon a large pretrained diffusion model such as BrainLDM\cite{pinaya2022brain_brainLDM}, VCM controls spatially fine-grained layouts from various new conditions. Using the generation abilities of diffusion models, VCM can versatilely perform various tasks in medical images such as (b-d) data synthesis with labels, (e) super-resolution, and (f) image translation. We notate the used conditions in blue colored boxes, including 1D scalars input for BrainLDM.}
    \label{fig:versatile_examples}
\end{center}%

}]

\begin{abstract}
   Spatial control methods using additional modules on pretrained diffusion models have gained attention for enabling conditional generation in natural images. These methods guide the generation process with new conditions while leveraging the capabilities of large models.
   They could be beneficial as training strategies in the context of 3D medical imaging, where training a diffusion model from scratch is challenging due to high computational costs and data scarcity.
   However, the potential application of spatial control methods with additional modules to 3D medical images has not yet been explored.
   In this paper, we present a tailored spatial control method for 3D medical images with a novel lightweight module,  Volumetric Conditioning Module (VCM).
   Our VCM employs an asymmetric U-Net architecture to effectively encode complex information from various levels of 3D conditions, providing detailed guidance in image synthesis.
   To examine the applicability of spatial control methods and the effectiveness of VCM for 3D medical data, we conduct experiments under single- and multimodal conditions scenarios across a wide range of dataset sizes, from extremely small datasets with 10 samples to large datasets with 500 samples. The experimental results show that the VCM is effective for conditional generation and efficient in terms of requiring less training data and computational resources. We further investigate the potential applications for our spatial control method through axial super-resolution for medical images. Our code is available at \url{https://github.com/Ahn-Ssu/VCM}
\end{abstract}

\section{Introduction}
\label{sec:intro}
Recently, \textbf{\textit{spatial control}} methods \cite{li2023gligen_T2Icontrol,zhang2023adding_controlNet,mou2023t2iadapter,avrahami2023spatext} have been proposed for conditional generation, guiding large pretrained diffusion models, such as Stable Diffusion \cite{rombach2022highresolution_LDM}, trained on massive data.
These spatial control methods employ additional lightweight modules that require small amounts of data to learn novel conditioning, demonstrating high image fidelity and alignment with a wide variety of given spatial input, such as semantic maps, bounding boxes, poses, sketches, layouts, and depth maps \cite{xue2023freestyle,voynov2022sketchguided_T2Icontrol,zheng2023layoutdiffusion}.
Another advantage of these spatial control methods is that only small modules need to be trained to control the reverse diffusion process with input conditions \cite{zheng2023layoutdiffusion,ham2023modulating_mcm,zhang2023adding_controlNet}.
Therefore, they can leverage the generation capabilities of pretrained diffusion models into downstream tasks in a computationally efficient way. 
In addition, training such small modules can alleviate the problems of poor generalizability, overfitting, and catastrophic forgetting caused by directly training large models \cite{serra2018overcoming_cata,hu2021lora_overfitting,ham2023modulating_mcm}.

The numerous benefits of these spatial control methods could be valuable for conditional generation in the 3D medical image domain, where publicly available training data is limited compared to natural images \cite{pinaya2023generative_monaiGen}, and the huge training costs of 3D diffusion models are usually unaffordable with an enterprise-level GPU.
Especially, their application in 3D medical images has the potential for downstream tasks, including super-resolution \cite{xu2024simultaneous_TFS_diff_SR} and image translation \cite{hyperGAE} with conditions of partial images and other imaging scans. Furthermore, such control methods can be utilized to synthesize high-quality, precise, and privacy-concern-free medical images with the same semantics as real patients, thus addressing the scarcity of public data in medical imaging \cite{Ktena2024_synImprove_nature,tian2023stablerep_synImprove_,azizi2023synthetic_synImprove_clf}. 
Nevertheless, the applicability of these spatial control approaches with additional modules \cite{zhang2023adding_controlNet,mou2023t2iadapter,ham2023modulating_mcm} suggested in 2D natural images has not been explored sufficiently to be used in 3D medical images. 
For instance, although there is potential for performance improvement when leveraging the generation ability of the pretrained models, most studies have trained the diffusion models from scratch \cite{xiang2023ddm2_imgEnhance,Fernandez_2022_synImprove_seg,Ktena2024_synImprove_nature,ozbey2023unsupervised_i2iTranslation}.
Therefore, we aim to adapt existing spatial control methods and propose a suitable network specifically designed for 3D medical images.

In this paper, we present a  \textit{\textbf{tailored spatial control}} method for 3D medical images using a novel network called the Volumetric Conditioning Module (VCM), which facilitates conditional generation by leveraging pretrained diffusion models. 
The proposed VCM learns spatial controls to guide the denoising process at each timestep while preserving the quality and capabilities of the large model by freezing its parameters.
To address the challenges of directly applying previous approaches from 2D natural images \cite{mou2023t2iadapter,ham2023modulating_mcm,zhang2023adding_controlNet} to volumetric spaces, we design our module as a lightweight and asymmetric U-Net architecture with time embedding that provides fine-grained guidance from various input conditions. For training in multimodal conditions, we employ a simple yet effective regularization method that maintains the generation quality of the large models in every conditioning case by applying dropout of the input modalities in the corresponding latent spaces. 


To examine the applicability of spatial control methods and the efficacy of VCM, we carry out experiments under single- and multimodal conditions across a wide range of dataset sizes, from extremely small datasets with 10 samples to large datasets with 500 samples.
Subsequently, we thoroughly compare and analyze the generation results with diverse aspects: 
\begin{itemize}
    \item
    We first investigate whether the spatial control methods used in 2D natural images can advance conditional generation in 3D medical image synthesis.
    Furthermore, we develop a novel lightweight module, Volumetric Conditioning Module (VCM), for spatial controls in 3D medical image generation.
    \item In contrast to natural images, most publicly available medical image datasets often consist of a few dozen data due to expensive 3D annotation and privacy concerns. Given this, we experiment with existing spatial control methods and our VCM across various sizes of datasets and different conditioning scenarios to delve into the optimal design corresponding to the available dataset size.
    \item It is important for the generated outputs to precisely align with the given conditions, since the image can be utilized in various applications of the medical image domain, such as data augmentation via semantics, super-resolution, and translation.
    For this reason, we measure the condition alignment of spatial control methods using the Dice score for segmentation masks and image difference for partial images. 
\end{itemize}

Furthermore, we apply our VCM to various applications of T1w brain MRI to suggest potential applications through spatial control methods in medical images (see \cref{fig:versatile_examples,subsec:application,supply:applications}). 
We hope that our careful design of VCM for spatial control and its results will help future research in applying control methods to various medical image applications.

\section{Related Work}
\label{sec:realWork}

\subsection{Controllable generation with diffusion models}
Although large Text-to-Image (T2I) diffusion models \cite{rombach2022highresolution_LDM,nichol2022glide,ramesh2022hierarchical_dallE,saharia2022photorealistic_imagen} have achieved promising results in conditional generation, there are challenges in obtaining the samples that accurately match the desired appearances, style, and composition using only 1D abstract conditions \cite{li2023gligen_T2Icontrol}. 
Therefore, numerous trial-and-error would be required to inspect the output and edit input prompts until the intended output is achieved. To address this issue, many controllable generation methods have been proposed for T2I diffusion models to synthesize images with specific conditions by subject-driven \cite{ruiz2023dreambooth_finetuning, gal2022image_textureInver}, style customizing \cite{sohn2023styledrop,chen2023artadapter}, advanced text conditions \cite{chefer2023attendandexcite_advText}, and drawing the region of interests \cite{rombach2022highresolution_LDM,Xie_2023_CVPR-smartbrush}.

\begin{figure}[t!]
\begin{center}
\includegraphics[width=0.9\linewidth]{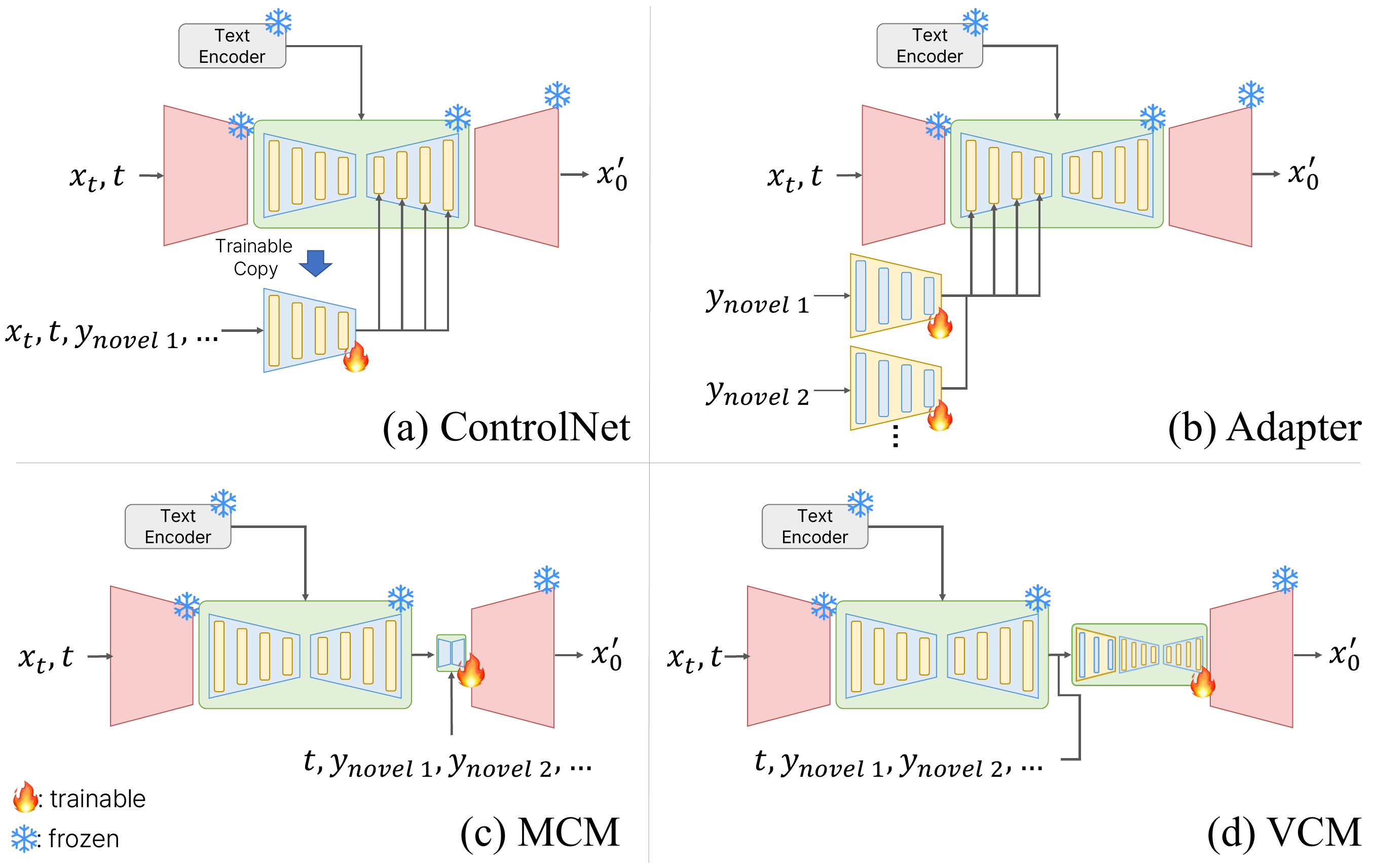}
\end{center} 
\caption{Comparison of controlling schemes. ControlNet \cite{zhang2023adding_controlNet} and T2I-Adpater \cite{mou2023t2iadapter} employ feature fusion to inject guidance. Meanwhile, MCM \cite{ham2023modulating_mcm}, remarkably lightweight module, and our VCM adapt a modulation approach.}
\label{fig:controlling_scheme}
\end{figure}

In particular, spatial control methods have been introduced to inject new localized controls into the denoising process by employing additional network modules, while locking the parameters of large diffusion models \cite{li2023gligen_T2Icontrol, ham2023modulating_mcm, voynov2022sketchguided_T2Icontrol,avrahami2023spatext,xue2023freestyle,voynov2022sketchguided_T2Icontrol,zheng2023layoutdiffusion}. 
ControlNet \cite{zhang2023adding_controlNet} and T2I-Adapter \cite{mou2023t2iadapter} are notable examples that provide novel spatial controls from various modalities for T2I pretrained diffusion models such as Stable Diffusion.
Both use the feature-fusion method to inject structure guidance into each output of the large model layers so that the additional module size corresponds to that of the large backbone models (\cref{fig:controlling_scheme} (a) and (b)).
MCM \cite{ham2023modulating_mcm}, another spatial control method, proposes the modulation approach for novel conditions (\cref{fig:controlling_scheme} (c)). MCM is a highly lightweight network, $\sim$0.4\% of Stable Diffusion, showing plausible results aligned with new conditions in terms of image quality and diversity.
Simple resizing of the condition images is used as a downsampling method for MCM to match the size of the latent spaces, serving as a compromise between accuracy and diversity in generation.

\subsection{Diffusion models in medical images}
\label{subsec:diffinMIA}
Building on the progress in computer vision, there has been a growing interest in diffusion models in the medical imaging field \cite{KAZEROUNI2023102846_medicalDiff_survey}. Extensive studies have been conducted for diffusion models in the medical domain, including image-to-image translation \cite{lyu2022conversion_mtDiffusion, ozbey2023unsupervised_i2iTranslation}, reconstruction \cite{G_ng_r_2023_adaDiff_recon, xie2022measurementconditioned_mcdiff_recon}, classification \cite{diffMIC_clsf}, super-resolution \cite{TMI2022diff_denoising_SR, wang2023inversesr, mao2023disc}, segmentation \cite{Fernandez_2022_synImprove_seg,kim2023diffusion_darl}, denoising \cite{TMI2022diff_denoising_SR}, anomaly detection \cite{anoDDIM,anoDDPM}, and generation \cite{pinaya2022brain_brainLDM, DISPR_medicalGeneration}. 
They have suggested novel approaches utilizing diffusion models in medical images, demonstrating superior performance compared to previous state-of-the-art performance of deterministic methods.

A publicly available large diffusion model, BrainLDM\cite{pinaya2022brain_brainLDM} is trained on the largest brain MRI dataset from UK Biobank \cite{Sudlow2015-fz__UKbioBank}.
As the model has learned the semantics of brain structures using given scalars paired with massive image data, it has the capabilities to generate T1w MRI scans based on input conditions with high-fidelity and diversity. 
Despite the rise of large diffusion models in medical imaging, few studies have explored their use to generate images under concrete conditions~\cite{wang2023inversesr}. 
As general approaches~\cite{zhang2023adding_controlNet,mou2023t2iadapter,ham2023modulating_mcm} in natural images have introduced well-generalized modules to guide with various spatial conditions to the frozen large foundation models, we propose pioneering Volumetric Conditioning Modules designed to control large diffusion models for 3D medical images. 
Our suggested module enables various conditional generation applications under concrete conditions, including volumetric semantic maps and partial images.

\section{Method}
We propose the Volumetric Conditioning Module (VCM), which enables the injection of controls into the reverse diffusion process using novel conditions originally unseen in the training of the large pretrained models. In this section, we briefly describe diffusion models and their preliminaries, and elaborate on our proposed methods, VCM.

\subsection{Preliminaries}
Diffusion models~\cite{ho2020denoising_ddpm,sohldickstein2015deep_1stDiffusion} are trained to generate data $x_0$ sampled from a data distribution.
To model a data distribution, the forward noising process for the input volume $x_0 \in \mathbb{R}^{1\times H\times W\times D}$ is defined with a specific variance schedule $\beta_1,...,\beta_T$ in $T$ timesteps:
\begin{equation}
    x_t=\sqrt{\bar{\alpha_t}}x_0 + \sqrt{1-\bar{\alpha_t}} \epsilon,
\label{eq:forward}
\end{equation}
where $\epsilon\sim N(0,I)$, $\alpha_t = 1 - \beta_t$ and $\bar{\alpha_t} = \Pi_{s=1}^{t}\alpha_s$.
In the reverse denoising process, a noise predictor $\epsilon_\theta(x_t,t)$ takes a noisy input $x_t$ and predicts the noise $\epsilon_t$ added in the forward process at timestep $t$. For diffusion models conditioned on a given set of conditions $y^*$, the reverse process is then $\epsilon_\theta(x_t,\tau(y^*), t)$ with the condition encoder $\tau(\cdot)$~\cite{rombach2022highresolution_LDM}. The reverse process is trained by optimizing the mean squared error between the noise prediction $\epsilon_t$ and the Gaussian noise $\epsilon \sim N(0,I)$.

Given a noised sample $x_t$ and the prediction $\epsilon_t = \epsilon_\theta(x_t, t)$, the denoised sample $x'_0$ can be approximated by a linear combination of $x_0$ and $\epsilon_t$~\cite{song2022denoising_ddim}:
\begin{equation}
    x_0' = \frac{x_t - \sqrt{1-\bar{\alpha_t}}\epsilon_t}{ \sqrt{\alpha_t}}
\label{eq:backward}
\end{equation}
and the $x_{t-1}$ in the next denoised timestep can be calculated using variational inference objective:
\begin{equation}
    x_{t-1} = \sqrt{\bar{\alpha}_{t-1}} x'_0 + \sqrt{1-\bar{\alpha}_{t-1}-\sigma_t^2}\epsilon_t + \sigma_t\epsilon_t
\label{eq:nexttime}
\end{equation}
where $\sigma_t = \eta \cdot \sqrt{(1-\bar{\alpha}_{t-1})/(1-\bar{\alpha}_t})\cdot\sqrt{(1-\bar{\alpha}_t)/\bar{\alpha}_{t-1}}$. If $\eta = 0$, \cref{eq:nexttime} becomes deterministic, enabling more efficient sampling within fewer timesteps than that of DDPM~\cite{ho2020denoising_ddpm}. 
In this study, we apply our approach to BrainLDM \cite{pinaya2022brain_brainLDM}, which is a publicly available 3D Latent Diffusion Model \cite{rombach2022highresolution_LDM,pinaya2023generative_monaiGen}. Since the calculations of the diffusion process are performed in latent spaces of the BrainLDM autoencoder, $x_t$ can be replaced with $z_t \in \mathbb{R}^{c\times h\times w\times d}$ from $z = E_\textit{AE}(x)$, and $x'_0$ can be replaced with $z'_0$ from $x_0' = D_\textit{AE}(z'_0)$ in \cref{eq:forward,eq:backward}.

\begin{figure}[!t]
\centerline{\includegraphics[width=0.9\columnwidth]{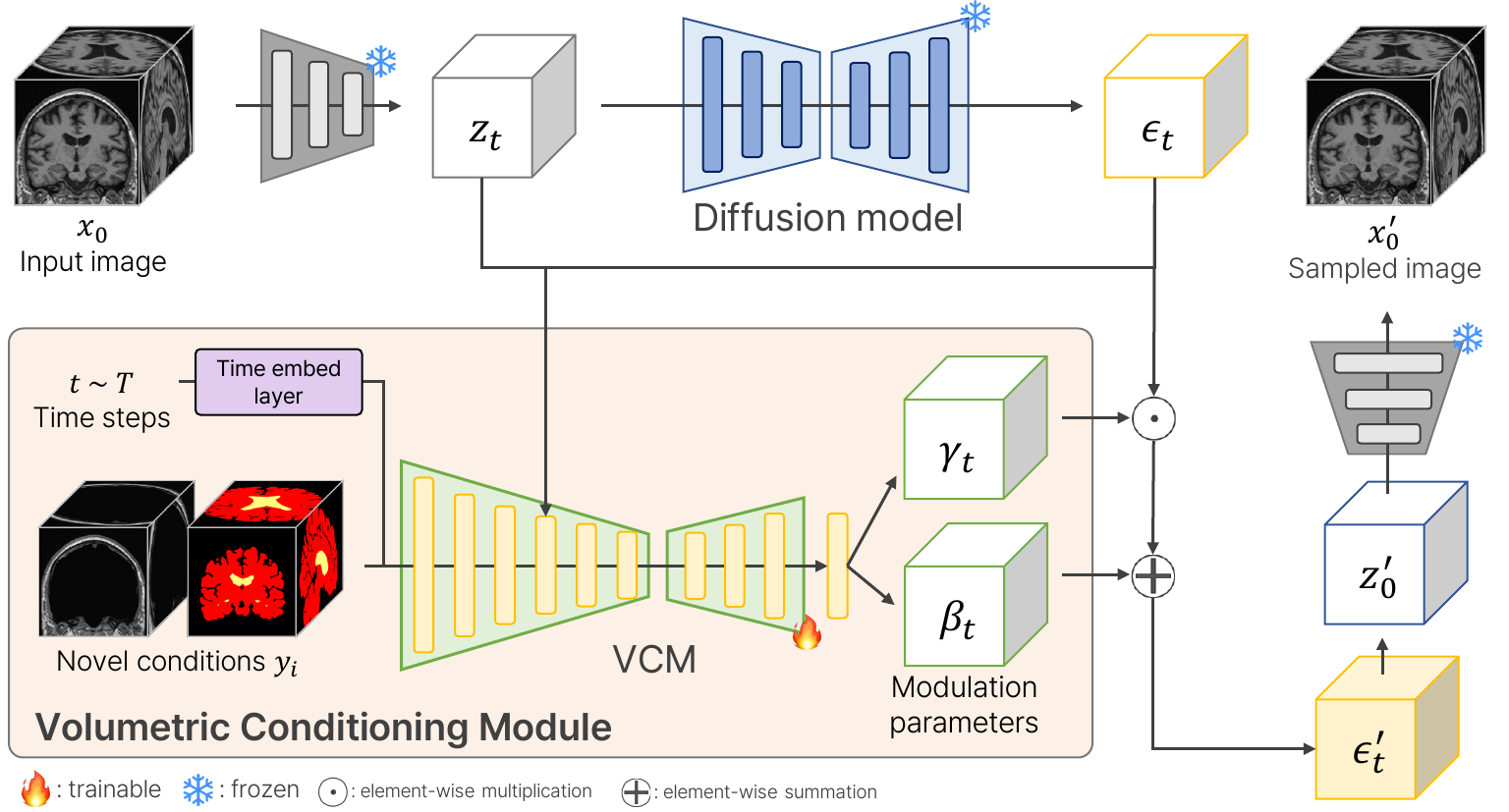}}
\caption{Illustration of the proposed VCM pipeline. VCM is located on the top of a pretrained diffusion model and modulates the model's output by leveraging the diffusion priors and new conditions as inputs. More details in \cref{supply:impl,fig:VCMdetails_modalitydrop}.}
\label{fig:VCM_pipeline}
\end{figure}

\subsection{Volumetric Conditioning Module}
\label{sec:VCM}
VCM injects additional conditions by modulating the output of the pretrained diffusion model at the top of the diffusion pipeline (see \cref{fig:controlling_scheme} (d)). 
VCM has an asymmetric U-Net architecture \cite{ronneberger2015unet} with time embedding layers, featuring a deeper encoder than the decoder. This design enables the precise encoding of complex information of spatial conditions from voxel spaces to latent spaces.
Due to its tailored design for medical images, VCM efficiently learns various spatial controls, from low-level volumetric representations (imaging modalities, partial images) to high-level ones (segmentation mask).

We visualize the VCM pipeline to modulate the output of the diffusion model at each timestep in \cref{fig:VCM_pipeline}. 
VCM can be integrated into the diffusion pipeline and fine-tuned for conditional generation while locking the large model's parameters.
To train VCM $\phi_\theta$, the input volume $x_0$ is fed into the autoencoder, producing latent $z_0 \in \mathbb{R}^{c\times h\times w\times d}$. 
Then, a timestep $t$ is randomly sampled and the noise is added to $z_0$ by the noise scheduler, resulting in $z_t$. 
The predicted noise $\epsilon_t$ is obtained by the LDM, $\epsilon_t = \epsilon_\theta(z_t, \tau(y^*), t)$. 
We refer to the input $z_t$ and the output $\epsilon_t$ as diffusion priors.
Given the set of spatial conditions $\{y_1, ..., y_n\}$ corresponding to $x_0$, the diffusion priors $\{z_t, \epsilon_t\}$ and the condition set $\{y_1, ... y_n\}$ are provided as the input for VCM with the timestep $t$, \ie, $\phi_\theta(z_t, \epsilon_t, y_1, ..., y_n, t)$. 
Afterward, VCM outputs two parameter tensors $\{\gamma_t, \beta_t\} \in \mathbb{R}^{c\times h\times w\times d}$, modulating the noise prediction $\epsilon_t$ as $ \epsilon_t' = \epsilon_t \odot (1+\gamma_t) \oplus \beta_t$. 
The modified prediction $\epsilon'_t$ is used to calculate the next denoised timestep by replacing $\epsilon_t$ with \cref{eq:nexttime}. 
Through the diffusion priors and the given set at each timestep, VCM learns the proper modulation for conditional guidance. 

For VCM, the objective function is to ensure that the training process captures the conditions of the input modalities. Through the non-Markovian process \cite{song2022denoising_ddim}, VCM can be trained by optimizing the mean squared error between the Gaussian noise $\epsilon$ and the modulated output $\epsilon_t'$ in the latent space of the autoencoder, \ie $\mathcal{L}_{\text{MSE}} = \text{MSE}(\epsilon, \epsilon_t')$.
In addition, we apply $L_1$ regularization to the VCM output parameters $\{\gamma_t, \beta_t\}$ to prevent VCM from drastically guiding the output only to align with the conditions, which could lead to performance degradation in the diffusion model.
Therefore, the final loss is as follows:

\begin{equation}
    \mathcal{L}_{\text{VCM}} = \mathcal{L}_{\text{MSE}} + \lambda{\mathcal{L}_1}(\gamma_t) + \lambda{\mathcal{L}_1}(\beta_t)
\label{eq:total_loss}
\end{equation}
where $\lambda$ is a scalar weighting term as $\lambda^{-1} = b \times c \times h \times w \times d$, $b$ is the batch size and $(h, w, d, c)$ are the dimensions of the latent space $z$. 


\textbf{Regularization for multimodal training}.
For effective training of VCM in multimodal synthesis, we use a regularization technique, called \textit{modality latent dropout}. 
We drop the output of the asymmetric encoder part by multiplying zero with probabilities $p_i$ corresponding to the condition modalities $y_i$. 
Our proposed method is simple, yet effective in preventing excessive reliance on a certain modality during training. 
This allows VCM to avoid unnecessary retraining after training at once, as VCM can condition with a subset of the trained modalities in a plug-and-play manner.
Furthermore, compared to the modality drop technique \cite{huang2022multimodal_modalDrop}, which is used in many multimodal syntheses, our \textit{modality latent dropout} makes VCM stable in training, preserving the quality of the generation of pretrained diffusion models.


\section{Experiments}
\label{sec:experiments}
In this section, we validate the spatial control methods used in 2D natural images and our VCM under single- and multimodal conditions across a wide range of training data sizes (10, 50, and 500 samples) on T1w brain MRI. All experimental configuration details and model implementations are described in \cref{supply:expr_details}.

\textbf{Datasets}. The data used in our study are a total of 604 healthy T1w Brain MRI scans obtained from the Alzheimer’s Disease Neuroimaging Initiative (ADNI) database~\cite{jack2008alzheimer_ADNI}. 
To acquire spatial conditions, we extract the segmentation masks of the lateral ventricle (LV) and brain using SynthSeg \cite{billot2023synthseg}.
The partial images of the skull were extracted by removing the brain region. 
Considering the data scarcity in medical images, we randomly separate the 604 MRI scans and acquired labels into three cases: 1) an extremely small dataset of 10 scans, 2) a small dataset of 50 scans, and 3) a large dataset of 500 scans for training. To measure the performance of conditional generation, the validation dataset comprises 44 scans, and its semantic mask and skull image are utilized to sample images.

\textbf{Evaluation Metrics}. For synthetic image evaluation, we employ
the Frechet Inception Distance (FID 2D) \cite{heusel2018gans_FID},  
Structural Similarity Index Measure (SSIM), 
Multi-Scale SSIM (MS-SSIM),
and Learned Perceptual Image Patch Similarity (LPIPS) \cite{zhang2018unreasonable_LPIPS} 
to assess image quality, structural similarities, and diversity, respectively. 
We also adopt the 3D Frechet Inception Distance (FID 3D) with pretrained MedicalNet \cite{chen2019med3d_medicalNet} for 3D medical images as applied in Med-DDPM \cite{Dorjsembe_2024_medDDPM}. 
To measure alignment with the given conditions, we use the Dice Similarity Coefficient (DSC), Hausdorff Distance (95\%, HD95), and Average Symmetric Surface Distance (ASSD) for segmentation masks. We also calculate the mean squared error between the real and synthesized data of skull images (skull-dist).
We report the average values of the validation dataset for each evaluation metric, highlighting the best performance in bold and the second-best performance with an underline.

\textbf{Implementation Details}. 
We employ BrainLDM \cite{pinaya2022brain_brainLDM}, creating a high-quality image of T1w brain MRI, as a backbone diffusion model for all spatial control methods.
We implement VCM as a time-conditioned U-Net \cite{ronneberger2015unet} having a deeper encoder than the decoder by modifying a diffusion network of MONAI \cite{pinaya2023generative_monaiGen}.
In VCM, the diffusion priors $\{z_t, \epsilon_t\}$ are aggregated with the latent outputs of the 3rd stage by simple concatenation. 
Lastly, we add a split head at the top of the last convolution layer of the VCM to produce the modulation parameters $\gamma_t$ and $\beta_t$ corresponding to the shift and the scale tensor, respectively.


\textbf{Comparison methods}. To investigate the applicability of spatial control methods and verify the efficacy of VCM, we employ other control methods of 2D natural images: ControlNet \cite{zhang2023adding_controlNet}, MCM \cite{ham2023modulating_mcm}, and T2I-Adapter \cite{mou2023t2iadapter}. For 3D medical image generation, they are re-implemented by substituting the 2D operations with the 3D counterparts referring to their description. 
In addition, we employ a large MCM (MCM-L) and lightweight variations of ControlNet (ControlNet-LITE and -MLP) to verify the efficacy of our design, emphasizing that its success is due to proper network design rather than simply the number of parameters compared to other methods.
To validate the effectiveness of the spatial control methods, 
we also test direct training a vanilla large diffusion model, LDM (from scratch), and fine-tuning of BrainLDM, LDM (fine-tuning).
Simultaneously, we report the results of the backbone diffusion model, BrainLDM, to show the controllability improvements and the quality preservation of spatial control methods.

\begin{table}[!t]
\caption{Evaluation result of alignment scores with spatial control via single condition, the lateral ventricle mask.}
\centering
\setlength{\tabcolsep}{5.3pt}
\fontsize{7}{7.5}\selectfont
\label{tab:LV-alignment}
\begin{tabular}{@{}c|c|c|ccc@{}}
\toprule
Method                              & \# params                & \# train data & HD 95 ↓         & DSC ↑         & ASSD ↓         \\ \midrule
\multirow{2}{*}{BrainLDM (text)} & \multirow{2}{*}{553.17M} & \multirow{2}{*}{-} & \multirow{2}{*}{20.154} & \multirow{2}{*}{0.594} & \multirow{2}{*}{2.658} \\
                                    &                          &               &                 &                &                \\ \midrule
\multirow{3}{*}{LDM (from scratch)} & \multirow{3}{*}{475.42M} & 10            & 33.892          & 0.656          & 17.076         \\
                                    &                          & 50            & 17.316          & 0.724          & 1.912          \\
                                    &                          & 500           & 15.473          & 0.840          & 1.090          \\ \midrule
\multirow{3}{*}{LDM (fine-tuning)}  & \multirow{3}{*}{553.18M} & 10            & 15.479          & 0.718          & 1.960          \\
                                    &                          & 50            & 18.995          & 0.732          & 1.787          \\
                                    &                          & 500           & 15.516          & 0.826          & 1.164          \\ \midrule
\multirow{3}{*}{MCM}                & \multirow{3}{*}{11.45M}  & 10            & 18.583          & 0.706          & 1.973          \\
                                    &                          & 50            & 17.221          & 0.690          & 2.045          \\
                                    &                          & 500           & 18.842          & 0.700          & 2.020          \\ \midrule
\multirow{3}{*}{MCM-L}              & \multirow{3}{*}{44.92M}  & 10            & 15.490          & 0.776          & 1.544          \\
                                    &                          & 50            & 17.141          & 0.768          & 1.561          \\
                                    &                          & 500           & 16.056          & 0.818          & 1.249          \\ \midrule
\multirow{3}{*}{T2I-adapter}        & \multirow{3}{*}{223.41M} & 10            & 15.767          & 0.803          & 1.372          \\
                                    &                          & 50            & 15.673          & 0.838          & 1.139          \\
                                    &                          & 500           & 16.231          & 0.873          & 0.909          \\ \midrule
\multirow{3}{*}{ControlNet}         & \multirow{3}{*}{193.88M} & 10            & \textbf{14.173} & {\ul 0.867}    & {\ul 0.961}    \\
                                    &                          & 50            & 16.453          & 0.868          & 0.939          \\
                                    &                          & 500           & \textbf{12.000} & \textbf{0.919} & \textbf{0.595}          \\ \midrule
\multirow{3}{*}{ControlNet-LITE}    & \multirow{3}{*}{63.56M}  & 10            & 17.612          & 0.786          & 1.420          \\
                                    &                          & 50            & \textbf{14.108} & 0.833          & 1.167          \\
                                    &                          & 500           & 14.506          & 0.917          & {\ul 0.609}    \\ \midrule
\multirow{3}{*}{ControlNet-MLP}     & \multirow{3}{*}{9.03M}   & 10            & 15.400          & 0.850          & 1.048          \\
                                    &                          & 50            & {\ul 14.961}    & {\ul 0.891}    & {\ul 0.796}    \\
                                    &                          & 500           &  13.709          & {\ul 0.918}    & \textbf{0.595} \\ \midrule
\multirow{3}{*}{VCM (Ours)}         & \multirow{3}{*}{45.19M}  & 10            & {\ul 14.872}    & \textbf{0.877} & \textbf{0.906} \\
                                    &                          & 50            & 15.776          & \textbf{0.893} & \textbf{0.789} \\
                                    &                          & 500           & {\ul 13.346}    & {\ul 0.918}    & 0.610          \\ \bottomrule
\end{tabular}
\end{table}

\subsection{Spatial controls via single condition}\label{sec:1st_exp}
We first investigate the applicability of spatial control methods in 3D medical images with LV masks as a single condition. 
The goal is to verify whether a pretrained diffusion model using the 1D scalar input of the LV volume can be manipulated to reflect the newly given volumetric LV mask condition through these methods.
The alignment performance of spatial control methods and direct training approaches is summarized in \cref{tab:LV-alignment}.

All spatial control methods present improvements in condition fidelity compared to direct training of large diffusion models.
Remarkably, for all datasets, our VCM consistently exhibits robust performance in terms of the size of training data.
In particular, on the extremely small and small datasets, VCM achieves the highest alignment scores in the DSC and ASSD, demonstrating data efficiency and generalizability in spatial controls.
ControlNet \cite{zhang2023adding_controlNet} shows the best alignment scores for all metrics when trained on the large dataset, verifying that it has the benefit of capturing complex structures.
On the large dataset, although the size of VCM is five times smaller than that of ControlNet, VCM achieves the most competitive performance in terms of the HD and DSC.
In contrast, with few training samples, the results of large spatial control modules such as ControlNet \cite{zhang2023adding_controlNet} and T2I-Adapter \cite{mou2023t2iadapter} show lower alignment scores than VCM under the DSC and ASSD. 
One possible explanation is that the generalizability issues arise due to the complexity of the module when training with few samples. 

To verify that the efficacy and robustness of VCM are due to its tailored architecture rather than only the small number of parameters, we analyze various spatial control methods by varying their complexity.
ControlNet-LITE and -MLP, lightweight variations of ControlNet \cite{zhang2023adding_controlNet}, display poorer performance than VCM with few training instances. 
Despite the smaller parameters, their alignment scores underperform compared to ControlNet for the extremely small dataset.
In contrast, both models achieve high scores comparable to ControlNet on the large dataset. This suggests that, with abundant training data, the ControlNet architecture provides sophisticated guidance regardless of its complexity.
Other small modules for spatial controls, MCM \cite{ham2023modulating_mcm} and MCM-L show the lowest performance among spatial control methods across all alignment metrics in every dataset. 
Its downsampling method is prone to information loss in volumetric spaces, making it difficult to achieve fine-detailed controls.
By comparing spatial control methods in varying complexity and architectures, we validate that our VCM offers advantages in data efficiency and performance due to its tailored design.

\begin{figure}[!t]
\centerline{\includegraphics[width=0.9\columnwidth]{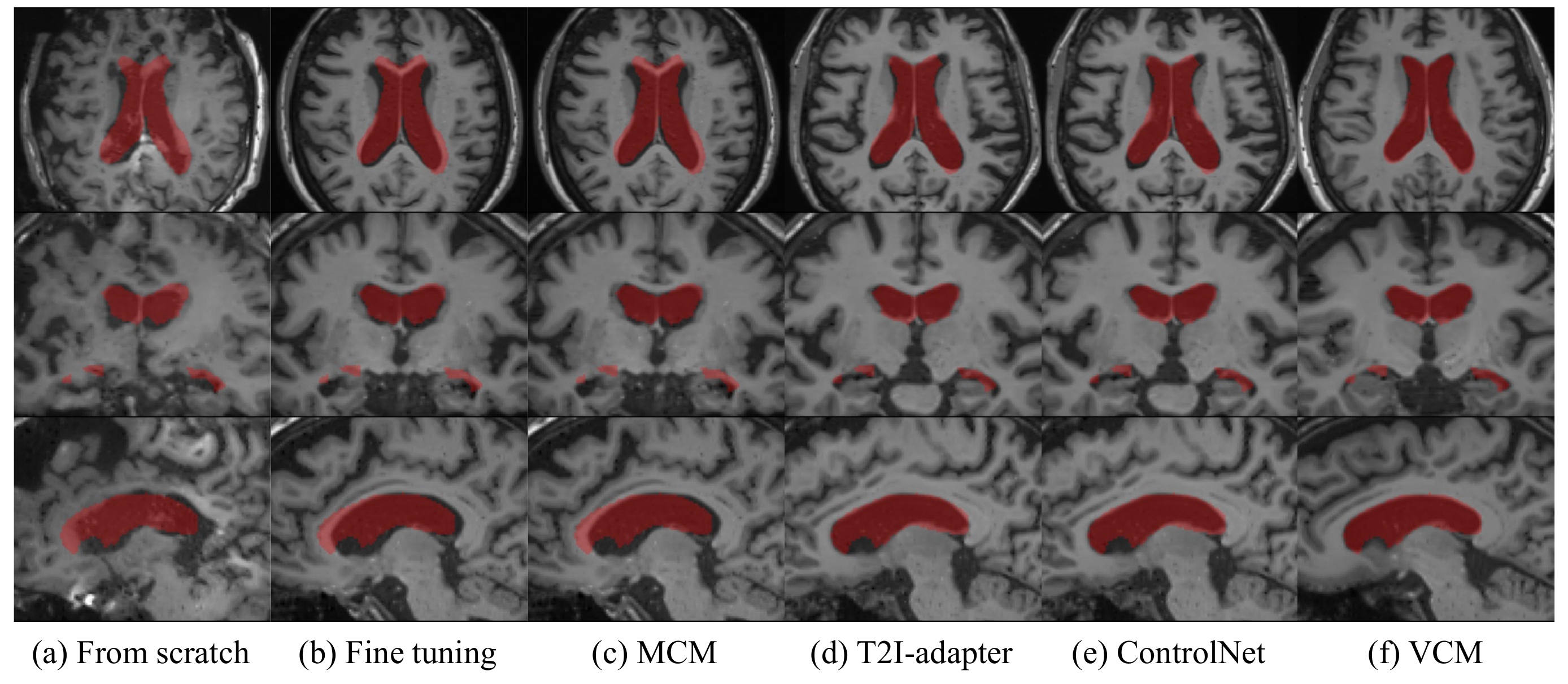}}
\caption{Synthetic images from various conditional generation methods trained with 50 training data with the LV mask condition.}
\label{fig:LV-overview}
\end{figure}

For qualitative evaluation, \cref{fig:LV-overview} depicts the sampled output of T1w brain MRIs through all spatial control methods and direct training approaches trained on the small dataset. Although all spatial control methods show high-quality synthesized images, only ControlNet (e) and our VCM (f) demonstrate precise condition alignment with the given LV mask (red color). However, direct training of diffusion models by fine-tuning or from scratch shows not only poor condition alignments, but also unsatisfactory image qualities.
The results highlight that employing spatial control methods is beneficial under a limited dataset while maintaining the quality of the generation.

In summary of the single condition experiments, we delve into the feasibility and applicability of various spatial control methods and demonstrate the superior efficiency and effectiveness of our VCM in condition alignment.


\begin{table*}[!t]
\centering
\caption{Evaluation results of alignment scores and image fidelity with spatial control methods under multimodal conditioning scenario.}
\label{tab:multi-expr}
\setlength{\tabcolsep}{5pt}
\fontsize{7}{7.5}\selectfont
\begin{tabular}{@{}c|c|c|cccc|ccccc@{}}
\toprule
Method                              & \# params                & \# train data      & HD 95 ↓                 & DSC ↑                 & ASSD ↓                 & skull-dist ↓           & MS-SSIM ↑              & SSIM ↑                 & FID 2D ↓                & FID 3D ↓               & LPIPS ↑                \\ \midrule
\multirow{2}{*}{BrainLDM (text)}     & \multirow{2}{*}{553.17M} & \multirow{2}{*}{-} & \multirow{2}{*}{18.407} & \multirow{2}{*}{0.710} & \multirow{2}{*}{2.383} & \multirow{2}{*}{0.029} & \multirow{2}{*}{0.521} & \multirow{2}{*}{0.434} & \multirow{2}{*}{84.511} & \multirow{2}{*}{2.247} & \multirow{2}{*}{0.340} \\
                                    &                          &                    &                         &                        &                        &                        &                        &                        &                         &                        &                        \\ \midrule
\multirow{3}{*}{LDM (from scratch)} & \multirow{3}{*}{475.43M} & 10                 & 95.595                  & 0.412                  & 70.595                 & 0.457                  & 0.598                  & 0.400                  & 92.142                  & 4.143                  & 0.492                  \\
                                    &                          & 50                 & 15.863                  & 0.828                  & 1.513                  & 0.025                  & 0.765                  & 0.537                  & 83.555                  & 2.695                  & 0.330                  \\
                                    &                          & 500                & 16.064                  & 0.863                  & 1.258                  & 0.013                  & 0.856                  & 0.641                  & 77.828                  & 1.451                  & 0.247                  \\ \midrule
\multirow{3}{*}{LDM (fine-tuning)}  & \multirow{3}{*}{553.19M} & 10                 & 18.089                  & 0.774                  & 1.953                  & 0.023                  & 0.679                  & 0.472                  & 88.020                  & 2.350                  & 0.390                  \\
                                    &                          & 50                 & 16.578                  & 0.823                  & 1.545                  & 0.017                  & 0.788                  & 0.564                  & 80.000                  & 1.632                  & 0.278                  \\
                                    &                          & 500                & 15.908                  & 0.884                  & 1.097                  & 0.013                  & 0.885                  & 0.663                  & 77.416                  & 1.939                  & 0.237                  \\ \midrule
\multirow{3}{*}{MCM}                & \multirow{3}{*}{11.45M}  & 10                 & 18.944                  & 0.684                  & 2.181                  & 0.027                  & 0.537                  & 0.435                  & 83.526                  & 3.258                  & 0.337                  \\
                                    &                          & 50                 & 17.966                  & 0.759                  & 2.034                  & 0.021                  & 0.648                  & 0.481                  & 82.456                  & 2.301                  & 0.312                  \\
                                    &                          & 500                & 17.127                  & 0.785                  & 1.803                  & 0.017                  & 0.705                  & 0.512                  & 83.050                  & 2.302                  & 0.302                  \\ \midrule
\multirow{3}{*}{MCM-L}              & \multirow{3}{*}{44.92M}  & 10                 & 17.529                  & 0.751                  & 2.197                  & 0.027                  & 0.560                  & 0.439                  & 84.434                  & 3.478                  & 0.337                  \\
                                    &                          & 50                 & 16.695                  & 0.804                  & 1.698                  & 0.019                  & 0.730                  & 0.517                  & 82.610                  & 2.335                  & 0.299                  \\
                                    &                          & 500                & 16.528                  & 0.836                  & 1.425                  & 0.016                  & 0.791                  & 0.565                  & 79.603                  & 2.162                  & 0.274                  \\ \midrule
\multirow{3}{*}{T2I-Adapter}        & \multirow{3}{*}{447.26M} & 10                 & 15.683                  & 0.837                  & 1.439                  & 0.016                  & 0.791                  & 0.585                  & 78.739                  & 2.069                  & 0.264                  \\
                                    &                          & 50                 & 15.869                  & 0.852                  & 1.325                  & 0.014                  & 0.821                  & 0.607                  & 78.036                  & 2.275                  & 0.252                  \\
                                    &                          & 500                & 15.544                  & 0.870                  & 1.209                  & 0.012                  & 0.860                  & 0.655                  & 76.993                  & 1.920                  & 0.242                  \\ \midrule
\multirow{3}{*}{ControlNet}         & \multirow{3}{*}{193.88M} & 10                 & 15.590                  & {\ul 0.878}            & {\ul 1.135}            & {\ul 0.012}            & {\ul 0.890}            & {\ul 0.684}            & {\ul 76.693}            & \textbf{1.349}         & 0.231                  \\
                                    &                          & 50                 & {\ul 15.079}            & {\ul 0.891}            & {\ul 1.028}            & \textbf{0.010}         & {\ul 0.903}            & {\ul 0.714}            & {\ul 75.255}            & {\ul 0.968}            & 0.218                  \\
                                    &                          & 500                & {\ul 13.899}            & {\ul 0.926}            & {\ul 0.728}            & \textbf{0.010}         & {\ul 0.943}            & {\ul 0.769}            & {\ul 72.643}            & {\ul 0.784}            & 0.196                  \\ \midrule
\multirow{3}{*}{ControlNet-LITE}       & \multirow{3}{*}{63.56M}  & 10                 & {\ul 15.363}            & 0.862                  & 1.251                  & 0.013                  & 0.854                  & 0.656                  & 77.277                  & 1.707                  & 0.236                  \\
                                    &                          & 50                 & 16.818                  & 0.878                  & 1.134                  & 0.011                  & 0.889                  & 0.685                  & 76.631                  & 1.432                  & 0.231                  \\
                                    &                          & 500                & 14.446                  & 0.909                  & 0.865                  & 0.011                  & 0.925                  & 0.744                  & 73.528                  & 1.230                  & 0.209                  \\ \midrule
\multirow{3}{*}{ControlNet-MLP}        & \multirow{3}{*}{9.03M}   & 10                 & 15.794                  & 0.855                  & 1.290                  & 0.013                  & 0.853                  & 0.656                  & 77.587                  & 2.160                  & 0.244                  \\
                                    &                          & 50                 & 16.296                  & 0.865                  & 1.205                  & 0.014                  & 0.869                  & 0.666                  & 76.092                  & 2.101                  & 0.237                  \\
                                    &                          & 500                & 14.499                  & 0.906                  & 0.880                  & 0.011                  & 0.923                  & 0.748                  & 73.686                  & 1.359                  & 0.206                  \\ \midrule
\multirow{4}{*}{VCM (Ours)}         & \multirow{4}{*}{45.88M}  & 10                 & \textbf{12.991}         & \textbf{0.921}         & \textbf{0.775}         & \textbf{0.010}         & \textbf{0.925}         & \textbf{0.746}         & \textbf{75.294}         & {\ul 1.429}            & 0.209                  \\
                                    &                          & 50                 & \textbf{14.521}         & \textbf{0.929}         & \textbf{0.710}         & {\ul 0.011}            & \textbf{0.936}         & \textbf{0.754}         & \textbf{73.427}         & \textbf{0.940}         & 0.202                  \\
                                    &                          & 500                & \textbf{12.932}         & \textbf{0.931}         & \textbf{0.684}         & {\ul 0.010}            & \textbf{0.943}         & \textbf{0.772}         & \textbf{71.551}         & \textbf{0.460}         & 0.193                  \\
                                    &                          & 3000               & 12.790                       & 0.931                      & 0.682                      & 0.010                      & 0.947                      & 0.777                      & 71.036                       & 0.462                      & 0.192                      \\ \bottomrule
\end{tabular}
\end{table*}

\subsection{Spatial control via multimodal conditions}\label{sec:2nd_exp}
In this section, we analyze spatial control methods under multimodal conditions of brain and LV segmentation masks and partial images of the skull. 
The purpose of this scenario is to evaluate how spatial control methods can precisely guide synthetic images close to real data from varying levels of novel input conditions (\ie, the semantic mask, and the skull images for high and low levels, respectively).
We summarize the qualitative evaluation of the alignment and generation quality for the control methods in \cref{tab:multi-expr}. 

Compared to \cref{sec:1st_exp}, our VCM outperforms the other methods in terms of the alignment scores and generation quality across every dataset, including the large dataset. In particular, there are still significant gaps in the DSC and ASSD compared to the suboptimal performance of ControlNet \cite{zhang2023adding_controlNet} in both the extremely small and the small datasets. 
The results show that VCM has a strong capability to precisely guide the generation process from various conditions. 
In addition to the alignment score results, VCM also demonstrates the best performance in terms of how the synthesized images are close to real images through the MS-SSIM, SSIM, FID 2D, and FID 3D, respectively. 
For VCM, we expand the training data up to 3000, nearly all healthy data in ADNI. Despite using six times more data than the large dataset, there are negligibly slight gains in alignment scores or image quality, indicating that VCM has high generalizability with training data.

Similarly in \cref{sec:1st_exp}, with few train samples, the large modules of ControlNet \cite{zhang2023adding_controlNet} and T2I-Adapter \cite{mou2023t2iadapter}  indicate lower performance than VCM. 
However, with abundant training samples, ControlNet is also one of the promising methods for spatial control due to its module capacity and the gradual increasing performance with the increasing amount of data. Nevertheless, T2I-Adapter, CNN modules to inject spatial controls, leaves a lot to be desired in the performance of both experiments. A conceivable analysis is that using time embedding layers in the additional modules such as ControlNet and VCM is beneficial to learn appropriate 3D guidance for the reverse diffusion process at each timestep. 
Furthermore, compared to \cref{sec:1st_exp}, the results of ContronlNet-LITE and -MLP are conspicuously degraded, revealing that it fails to handle various conditions.
The results of \cref{tab:multi-expr} show that our asymmetric architecture of VCM is robust to various levels of input with data efficiency and efficacy for conditional generation in 3D medical images.



\begin{figure}[!t]
\centerline{\includegraphics[width=1\columnwidth]{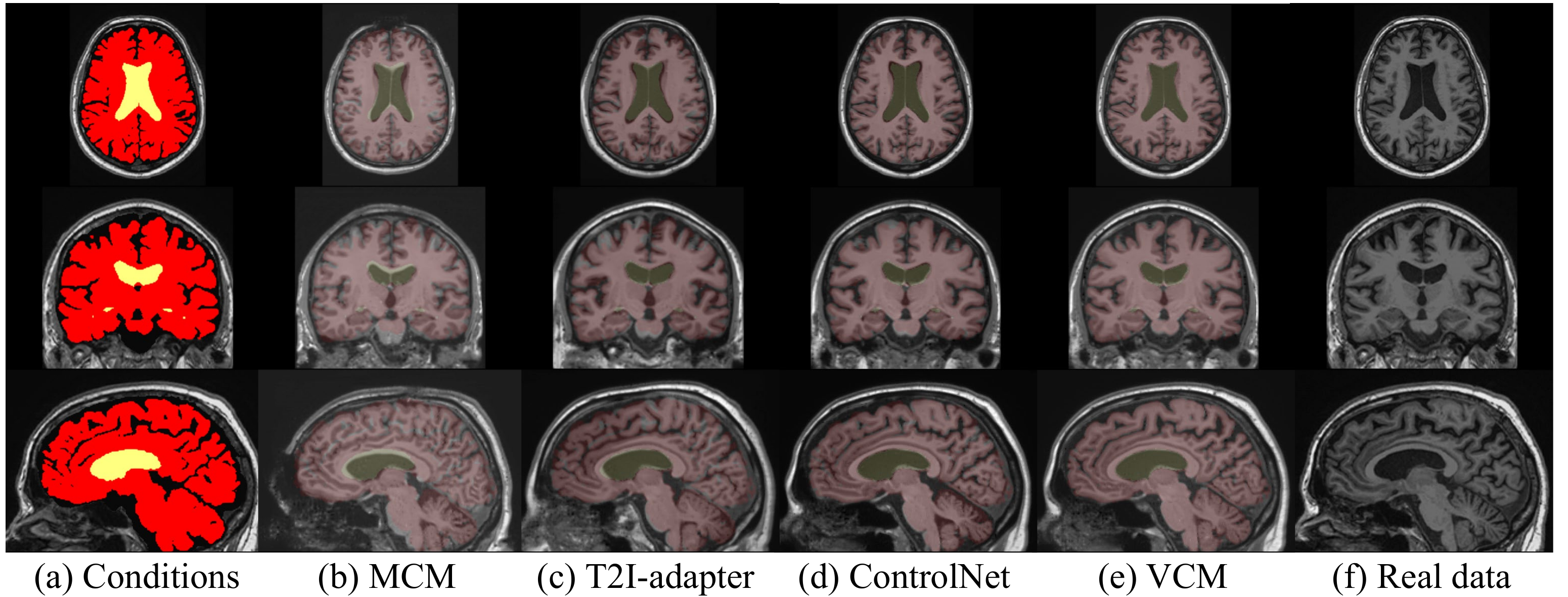}}
\caption{Synthetic images from various spatial control methods trained with 50 training data with multimodal conditions.}
\label{fig:3modal-overview}
\end{figure}

\begin{figure}[!t]
\centerline{\includegraphics[width=0.9\columnwidth]{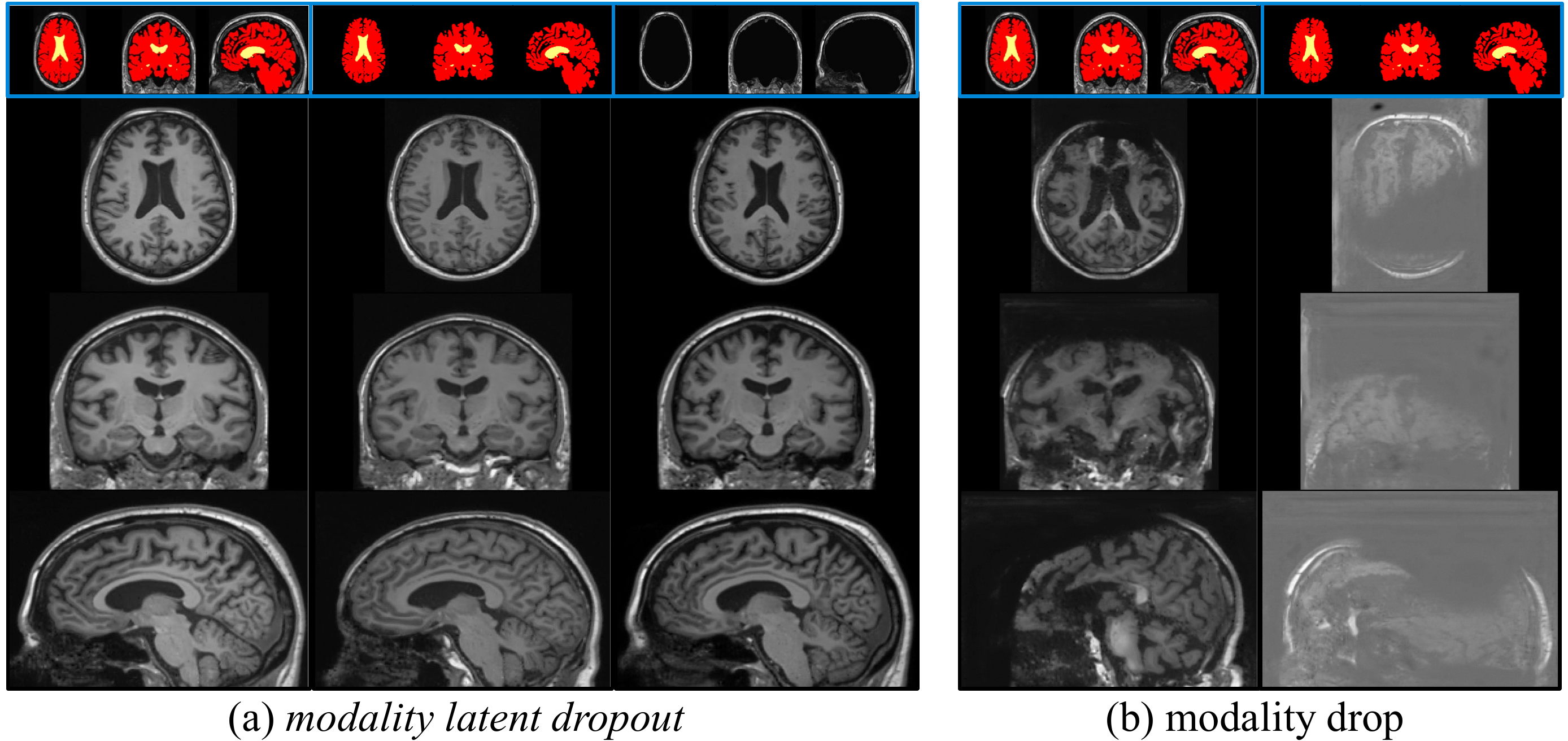}}
\caption{Comparison of sampling results between models trained with (a) the \textit{modality latent dropout} and (b) modality drop \cite{huang2022multimodal_modalDrop}.}
\label{fig:latentDrop_samples}
\end{figure}

\textbf{Regularization for multimodal training}. The \textit{modality latent dropout} has the same advantages as modality drop \cite{huang2022multimodal_modalDrop} in preventing the model from being overly dependent on specific conditions in multimodal synthesis. 
We test with both regularization techniques in 3D medical image generation. During training, an observation is that the \textit{modality latent drop} has the advantage over modality drop in that it can preserve the quality of the pre-trained diffusion model (see \cref{fig:latentDrop_samples}). VCM trained using the \textit{modality latent dropout} produces clear results, whereas the model learned with modality drop fails to synthesize normal T1w brain MRIs even after multiple sampling attempts. When the \textit{modality latent dropout} drops off a condition latent, it allows the VCM to learn only the remaining latent features during training under multimodal conditions in a way of plug-and-play.

\textbf{Ablation study}. We conduct an ablation study to determine which downsampling and composition for VCM yield the best performance. The dataset and conditions are the same configurations as in the \cref{sec:2nd_exp}. The results of the ablation study with 50 training samples are described in \cref{tab:ablation}. The trilinear refers to scaling the input conditions with trilinear interpolation without any encoder or time embedding layer in VCM. The simple CNN refers to the \textit{input hint} CNN encoder of ControlNet \cite{zhang2023adding_controlNet}, whereas the asymmetric encoder indicates an asymmetric UNet with a deeper encoder than the decoder. Our method employs a time embedding layer in an asymmetric UNet encoder. 

Compared to the trilinear for VCM, which shows the worst performance, using the simple CNN to encode the input conditions significantly improves overall performance. The asymmetric encoder enhances both condition alignments and generation qualities. In addition to the asymmetric encoder, incorporating time embedding layers in VCM improves stability, as shown by a reduction in the HD and comprehensive improvements. 
This ablation study supports our argument in \cref{sec:2nd_exp} that using time embedding layers for spatial controls is helpful to learn proper guidance at each timestep of the diffusion process.


\begin{table}[!t]
\centering 
\caption{Ablation study of network composition for VCM.}
\setlength{\tabcolsep}{5pt}
\fontsize{7}{7.5}\selectfont
\label{tab:ablation}
\begin{tabular}{@{}c|cccc@{}}
\toprule
Method       & HD 95 ↓         & DSC ↑          & ASSD ↓         & skull-dist ↓   \\ \midrule
trilinear    & 16.695          & 0.804           & 1.698          & 0.019          \\
simple CNN   & 15.586          & 0.919           & 0.810          & 0.011          \\
asymmetric encoder & 14.052          & 0.928           & 0.716          & \textbf{0.010} \\
+ time layer (Ours)         & \textbf{12.315} & \textbf{0.929}  & \textbf{0.710} & 0.011          \\ \midrule
Method       & MS-SSIM ↑       & FID 2D ↓        & FID 3D ↓       & LPIPS ↑        \\ \midrule
trilinear    & 0.730           & 82.610          & 2.335          & 0.299          \\
simple CNN   & 0.922           & 75.491          & 0.978          & 0.221          \\
asymmetric encoder & 0.934           & 74.505          & 0.972          & 0.207          \\
+ time layer (Ours)         & \textbf{0.936}  & \textbf{73.427} & \textbf{0.940} & 0.202          \\ \bottomrule
\end{tabular}
\end{table}

\subsection{Super-resolution}
\label{subsec:application}
In addition to precise image synthesis with single- and multimodal conditions, we further investigate the potential applications of spatial control methods through axial super-resolution.
Super-resolution (SR) is an important technique in MRI due to the trade-off between scanning time and image resolution. 
Spatial control methods can be a solution to generate high-quality MRIs given low-resolution (LR) image conditions. 
For the SR task, we converted 1 mm$^3$ T1w MRI scans into the LR scans with 4 mm (low sparsity) and 8 mm (high sparsity, thus challenging) resolution in the axial slices. 
Through these two different conditions, we validate the applicability and capability of our spatial control method in the SR scenarios.

\begin{figure}[!t]
\centerline{\includegraphics[width=0.9\columnwidth]{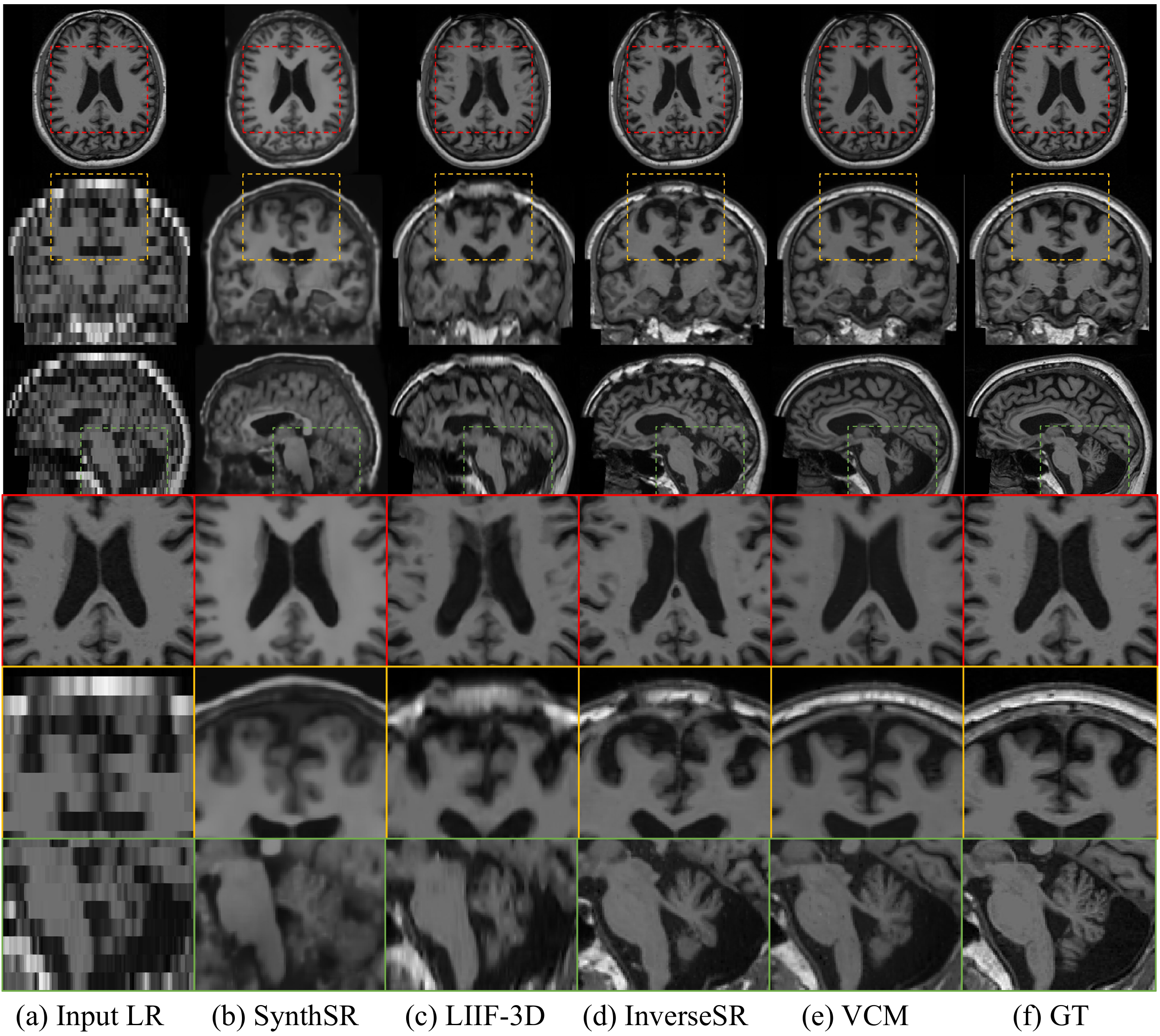}}
\caption{Qualitative results of 8 mm axial super-resolution.}
\label{fig:SRexample}
\end{figure}

To train VCM, we follow the same configurations as \cref{sec:1st_exp} and use 50 training samples. 
For comparisons, we employ various SR methods: 1) SynthSR \cite{iglesias2021joint_synthSR}, a pretrained model for brain MRI, 2) LIIF-3D \cite{chen2021learning_LIIF}, a trainable method that has shown state-of-the-art performance, and 3) InverseSR \cite{wang2023inversesr}, an optimization approach in voxel space for brain MRIs using a foundation model without training. For LIIF-3D, we train on the same 50 samples as VCM. The qualitative results are presented in \cref{tab:SR-eval}, whereas \cref{fig:SRexample,fig:SR_appendix} illustrate the quantitative results. 

In the low sparsity of 4 mm SR, VCM shows comparable performances with InverseSR \cite{wang2023inversesr}, an optimization-based approach in volumetric spaces. However, InverseSR, which uses the same backbone model \cite{pinaya2022brain_brainLDM} as ours, requires computational costs such as a huge GPU memory (\ie, 80 GB VRAM) and time for prediction due to iterative optimization. In contrast, VCM, a spatial control method, only takes less than 24 GB GPU memory for training and prediction. Nevertheless, LIIF-3D \cite{chen2021learning_LIIF} outperforms with significantly large margins in all metrics compared to the other methods. In the case of weakly degraded LR images, edge-cutting methods can achieve superior performance without leveraging the ability of large pretrained diffusion models.

Conversely, in the severely degraded LR images of 8 mm, where detailed features such as folds are hard to discern, VCM shows outstanding performance in all metrics, while LIIF-3D shows suboptimal performance. In addition, the result of InverseSR indicates worse scores than VCM.
In \cref{fig:SRexample}, the controlled image by (e) VCM exhibits the structures and details closest to (f) the ground truth. However, (c) LIIF-3D and (d) InverseSR produce brain MRI scans with abnormal appearances such as the destroyed skull and irregular LV shapes. SynthSR constructs the plausible LV shape in (b), while the other details such as the brain shapes and skull thickness are different from the ground truth. 
VCM captures the proper guidance from the sparse LR while the preserved ability of the pretrained diffusion model generates the condition-aligned SR images. 
Hence, spatial control methods could be beneficial in treating challenging medical image tasks.



\begin{table}[!t]
\centering
\fontsize{7}{7.5}\selectfont
\caption{Quantitative results of axial super-resolution.}
\label{tab:SR-eval}
\begin{tabular}{@{}c|c|ccc@{}}
\toprule
SR Task      & Method    & MAE                  & SSIM ↑               & PSNR ↑               \\ \midrule
\multirow{5}{*}{4 mm (x4)} & Cubic     & 0.031                & 0.794                & 23.165               \\
                     & SynthSR \cite{iglesias2021joint_synthSR}  & 0.055                & 0.662                & 19.222               \\
                     & LIIF-3D \cite{chen2021learning_LIIF}     & \textbf{0.011}                & \textbf{0.953}                & \textbf{32.401}               \\
                     & InverseSR \cite{wang2023inversesr} & {\ul0.020}                & {\ul 0.879}                & 27.509               \\
                     & VCM (Ours)       & 0.020                & 0.871                & {\ul 27.538}               \\ \midrule
\multirow{5}{*}{8 mm (x8)} & Cubic     & 0.053                & 0.610                & 19.780               \\
                     & SynthSR \cite{iglesias2021joint_synthSR}  & 0.057                & 0.623                & 19.020               \\
                     & LIIF-3D \cite{chen2021learning_LIIF}     & {\ul 0.026}                & {\ul 0.812}                & {\ul 25.121}               \\
                     & InverseSR \cite{wang2023inversesr} & 0.030                & 0.787                & 23.655               \\
                     & VCM (Ours)       & \textbf{0.023}                & \textbf{0.857}                & \textbf{27.518}               \\ \bottomrule
\end{tabular}
\end{table}

\section{Conclusion}
We introduce a novel spatial control method to leverage pretrained diffusion models for 3D medical images using a lightweight network called the Volumetric Conditioning Module (VCM). 
Through extensive experiments, we pioneeringly delve into the applicability of spatial control methods in 3D medical images, and show that our VCM effectively guides the large model with new conditions while demonstrating the efficiency in both parameters and data.
In addition, we explore the various applications of spatial control methods in 3D medical images. 
We anticipate that our VCM, which is feasible with an enterprise-level GPU with 24GB VRAM, will be widely adopted for various downstream tasks in medical imaging.
\clearpage
{\small
\bibliographystyle{ieee_fullname}
\bibliography{main}
}

\clearpage

\appendix


\twocolumn[{%
\begin{center}{\huge{\textbf{Appendix}}}\end{center}
\renewcommand\twocolumn[1][]{#1}%
\begin{center}
    \centering
    \captionsetup{type=figure}
    \includegraphics[width=0.9\textwidth]{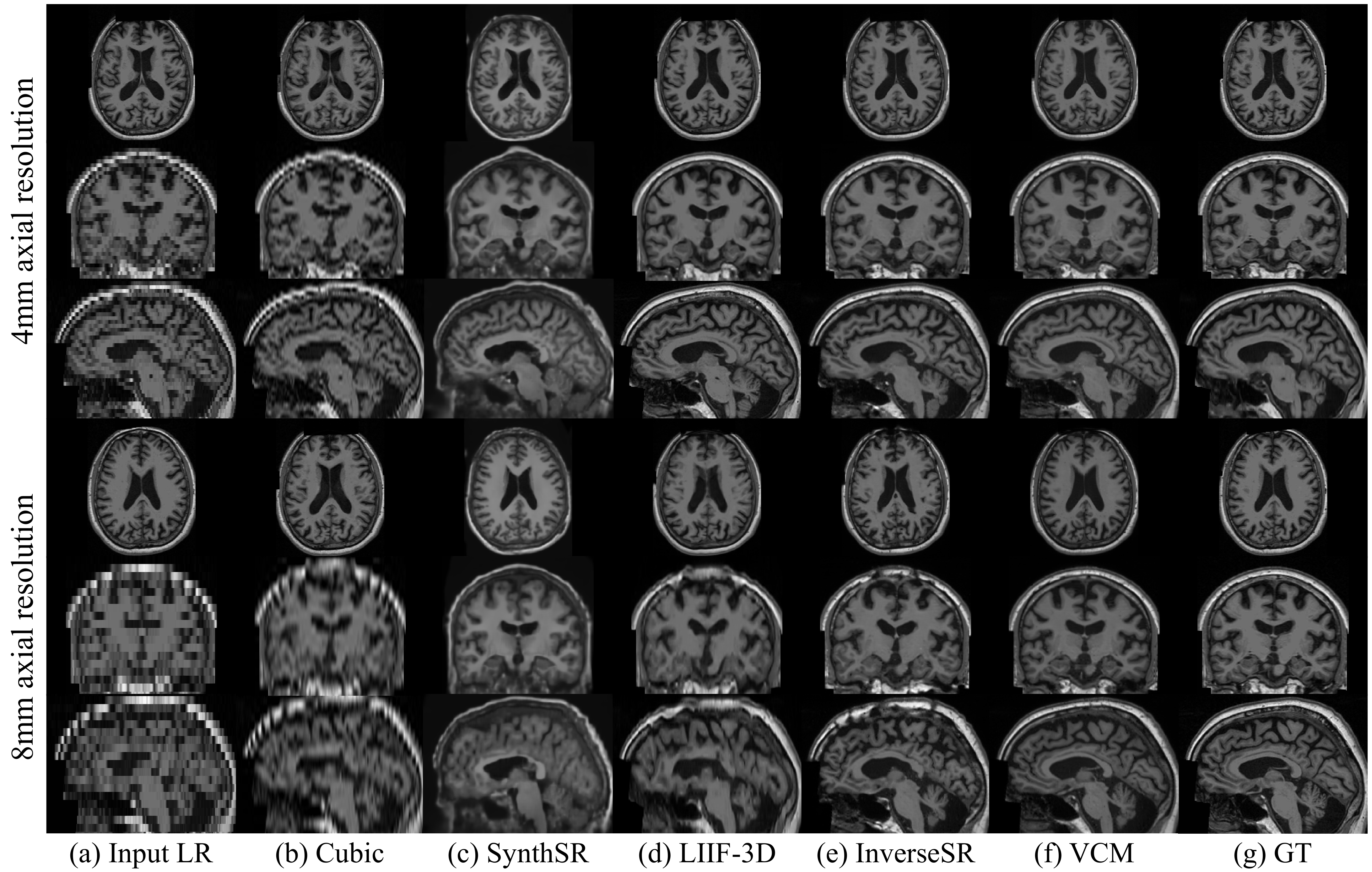}
    \captionof{figure}{Qualitative results of 4 mm and 8 mm axial super-resolution from \cref{subsec:application,fig:SRexample}.}
    \label{fig:SR_appendix}
\end{center}%
}]
\section{Additional experimental results}
\label{supply:applications}
We provide the more qualitative results from \cref{subsec:application} of the SR application and additional experiments and analysis with the LV masks 
Furthermore, as aforementioned in \cref{sec:intro,subsec:diffinMIA}, we here suggest certain usage cases of spatial control methods in medical images such as image translation and precise synthesis to elaborate on the answers to the following questions: \textit{``why are conditional generation approaches needed in the medical image domain?"} and \textit{``what are the applications of spatial control methods for medical images?"}.

\subsection{Super-resolution}
\label{subsupply:SR}

For better comparability between VCM and the other methods for the super-resolution (SR) in \cref{subsec:application,tab:SR-eval}, we visualize the more qualitative results of all methods. 

In the low sparsity of 4 mm SR, as LIIF-3D \cite{chen2021learning_LIIF} significantly outperforms in the metrics in \cref{tab:SR-eval}, the model produces the most similar images with the ground truth. For example, the image appearance of LIIF-3D has not only correct semantics in the brain structures (\eg, the lateral ventricles, the skull and white matter), but also high-frequency details such as brain folds (\eg, the gray matter). The optimized outputs of InverseSR \cite{wang2023inversesr} illustrate comparable visual quality with that of VCM. Since VCM learns spatial controls for the 4 mm SR task with only 50 training samples, the results are promising in both data and computational efficiency. However, compared to LIIF-3D, InverseSR and our VCM show poor quality in the details of brain MRI scans. 

In the higher sparsity of 8 mm SR, only VCM produces a normal brain MRI scan maintaining the shape of brain architectures such as the skull and the lateral ventricles, skull, and brain matters. The competitor in the 4 mm SR in both qualitative and quantitative results of \cref{tab:SR-eval,fig:SR_appendix}, InverseSR \cite{wang2023inversesr} shows some destroyed appearances in the generated output. In addition, LIIF-3D also creates an abnormal image, failing to perform the 8 mm SR. On the other hand, SynthSR \cite{iglesias2021joint_synthSR}, a pretrained model for brain MRI SR, shows consistent results regardless of the degradation of the input image. However, the method produces the poor metric scores and the mispredicted outputs such as the skull thickness and brain structure details.




\subsection{Image translation}
\label{subsupply:translation}
\begin{figure}[!t]
\centerline{\includegraphics[width=1\columnwidth]{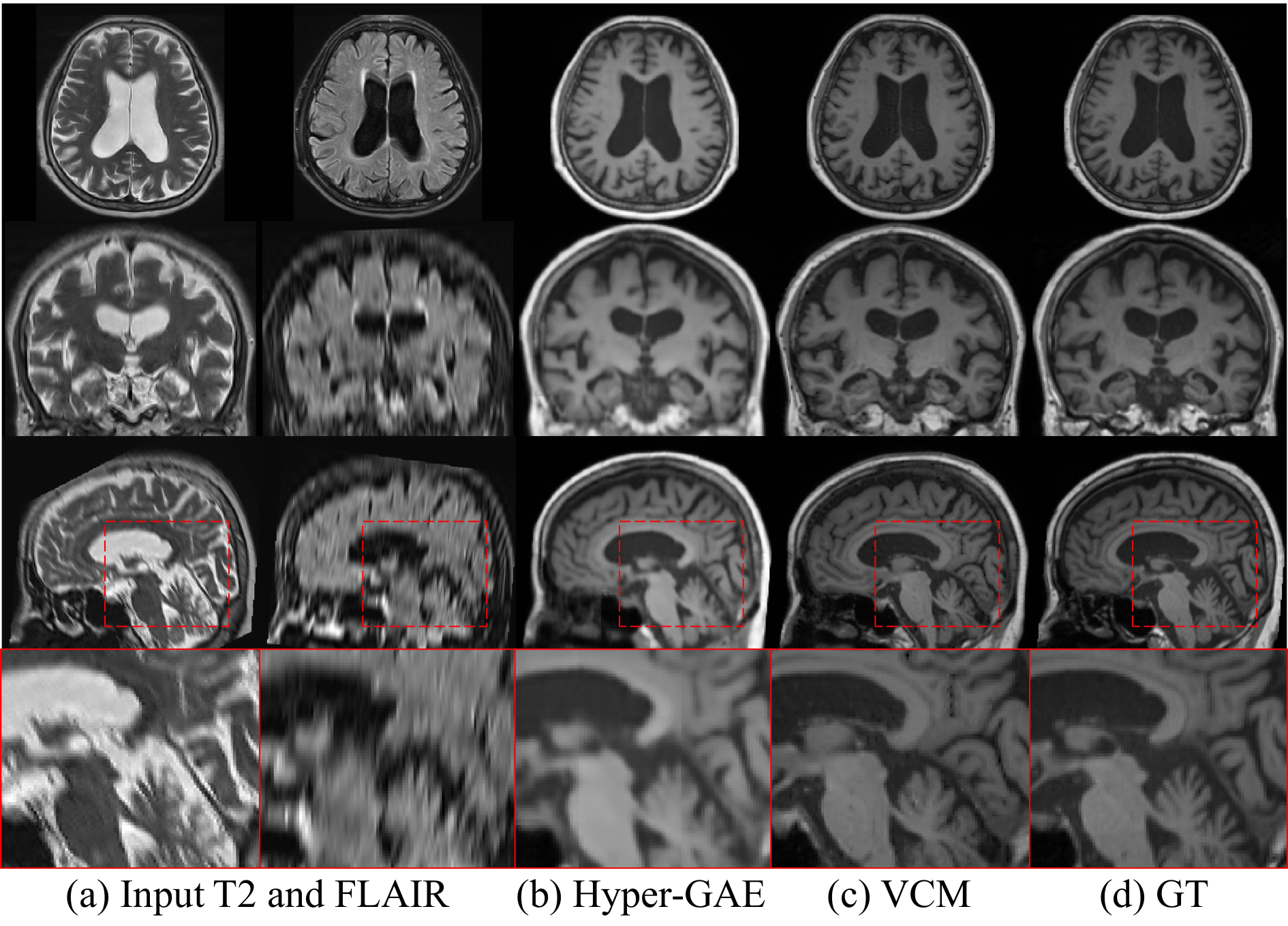}}
\caption{Comparison of generated T1w brain MRI results between Hyper-GAE \cite{hyperGAE} and our VCM for Image-to-Image translation. Used input conditions are T2 and FLAIR MRI scans.}
\label{fig:translation}
\end{figure}
Using the intrinsic magnetic resonance properties of tissues, we can acquire various MRI sequences such as T1, T2, and FLAIR, each offering exclusive information. A complete set of these sequences is ideal for an accurate diagnosis, but acquiring them in practice is often challenging due to the extended scanning time. Consequently, many studies have focused on predicting missing sequences from the available ones\cite{hyperGAE}.

In the context of image to image translation tasks in medical images, VCM can guide the diffusion process to synthesize the missing MRI sequence using the other sequences as spatial controls. In this experiment, we used T2 and FLAIR MRI scans as conditions to generate a translated T1w MRI scan. VCM was trained with 50 training sets of T2, FLAIR, and T1w brain MRI sequences.

As illustrated in \cref{fig:translation}, VCM synthesizes a T1w MRI scan, reflecting detailed features such as folds from the T2 and FLAIR scans.
Compared to HyperGAE \cite{hyperGAE}, the state-of-the-art 3D-based translation method, our VCM produces clearer synthetic images leveraging the pretrained diffusion model.
In addition, due to the high computational cost of 3D medical images, many 3D-based deep learning architectures employ patch-based scheme, which are prone to grid artifacts, as seen in the zoomed images of HyperGAE method (\cref{fig:translation} (b)).
In contrast, our VCM is free from grid artifacts, as it can handle the entire image even with an enterprise-level GPU (\eg, 24GB VRAM).

\subsection{Synthesis with precise semantics}
\label{subsupply:preciseSyn}

\begin{figure}[!t]
\centerline{\includegraphics[width=1\columnwidth]{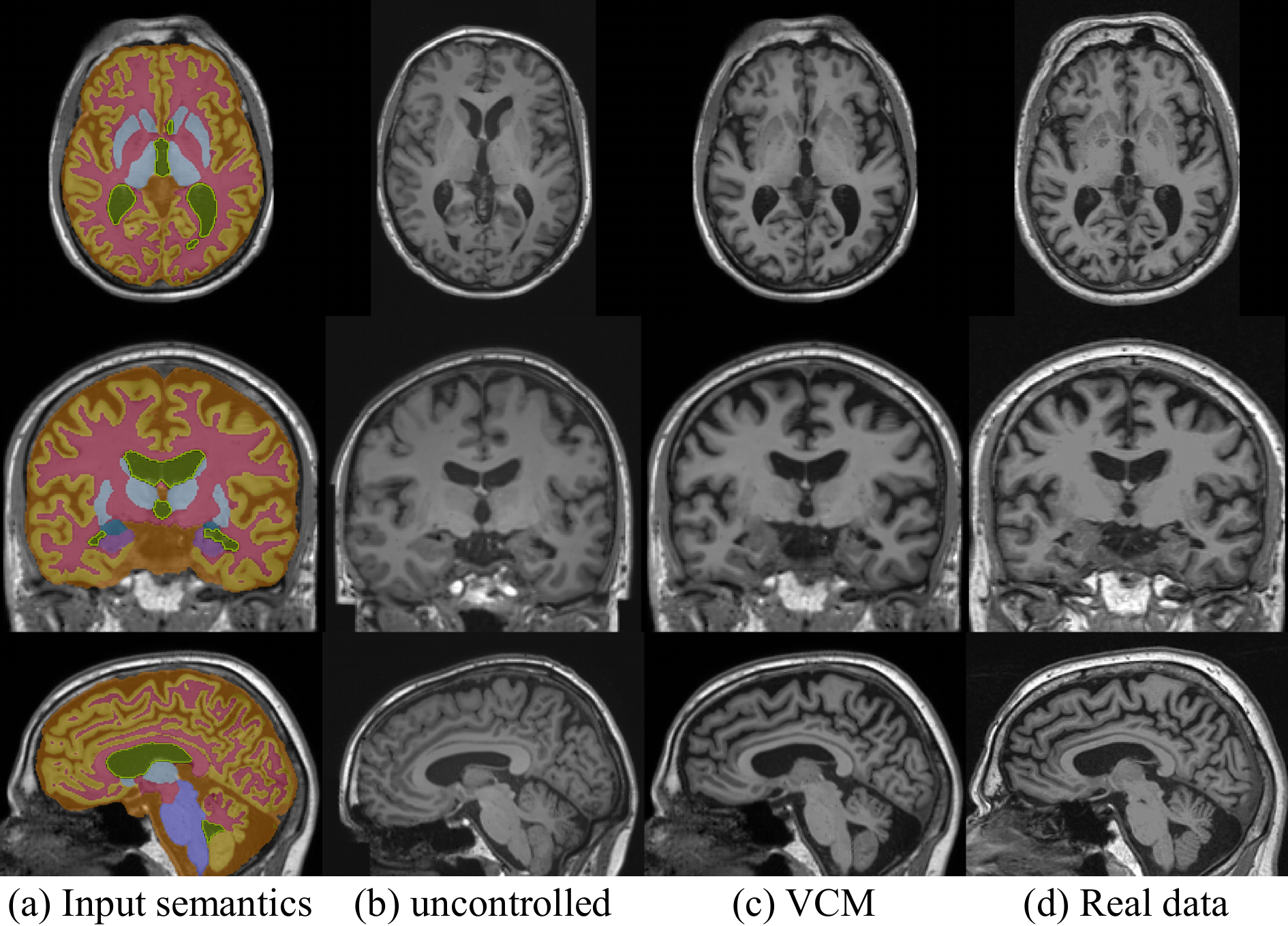}}
\caption{Samples of T1w brain MRI scans for uncontrolled output and controlled by our VCM with complex semantics.}
\label{fig:semantics-t1}
\end{figure}

\begin{figure}[!b]
\centerline{\includegraphics[width=0.9\columnwidth]{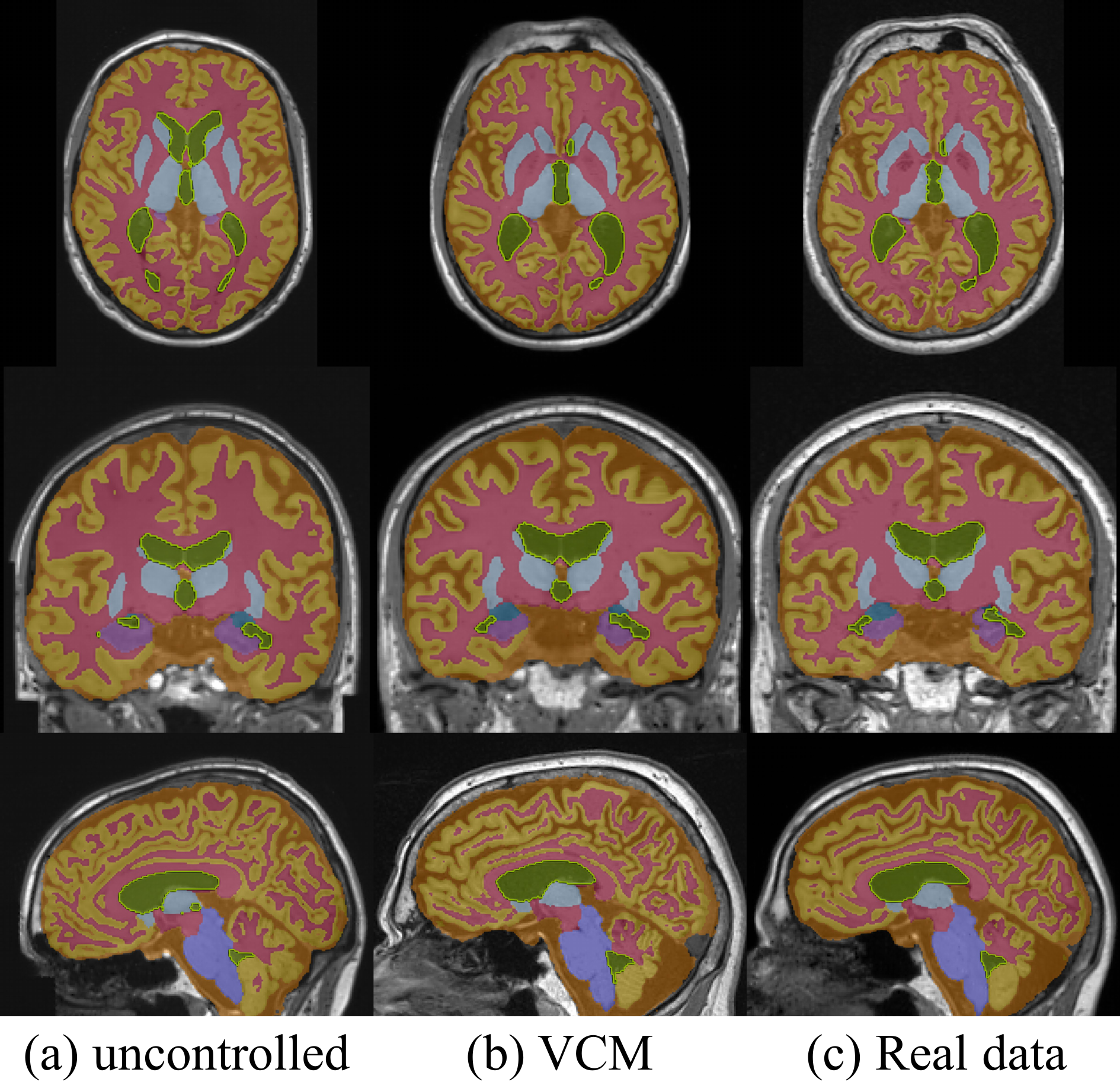}}
\caption{Comparison of the semantics of generated images between uncontrolled output and controlled by our VCM.}
\label{fig:semantics-seg}
\end{figure}


As suggested in \cref{sec:intro}, the scarcity of available data in medical images can be addressed using high-quality synthetic data that are privacy-concern-free and precisely match the anatomical composition of real patient data.
To generate these synthetic images, spatial control methods with the semantics of real patients can be a solution while also avoiding huge computational costs. 
To examine, we train our VCM with segmentation masks more complicated than those in \cref{sec:1st_exp}. 
Specifically, we increase the number of semantics classes to include: 1) cerebrospinal fluid, 2) white matter, 3) gray matter, 4) all ventricles, 5) thalamus, caudate, putamen, and accumbens, 6) amygdala, 7) hippocampus, and 8) brain stem. 

We visualize the samples generated by (a) BrainLDM \cite{pinaya2022brain_brainLDM} as uncontrolled and (b) controlled by our VCM with the semantics in \cref{fig:semantics-t1,fig:semantics-seg}. With the given input conditions, our VCM controls and generates the outputs of T1w brain MRI, precisely close to the real data, which is used to obtain the input semantics. 
Especially, not only the global appearance of the MRI scans (see \cref{fig:semantics-t1}), but also the generated semantic masks are highly homogeneous with the real data and the given condition (see \cref{fig:semantics-seg}). 
\subsection{Further analysis of spatial controls}
\label{subsupply:LV+extra}
\begin{figure}[b]
\centerline{\includegraphics[width=1.0\columnwidth]{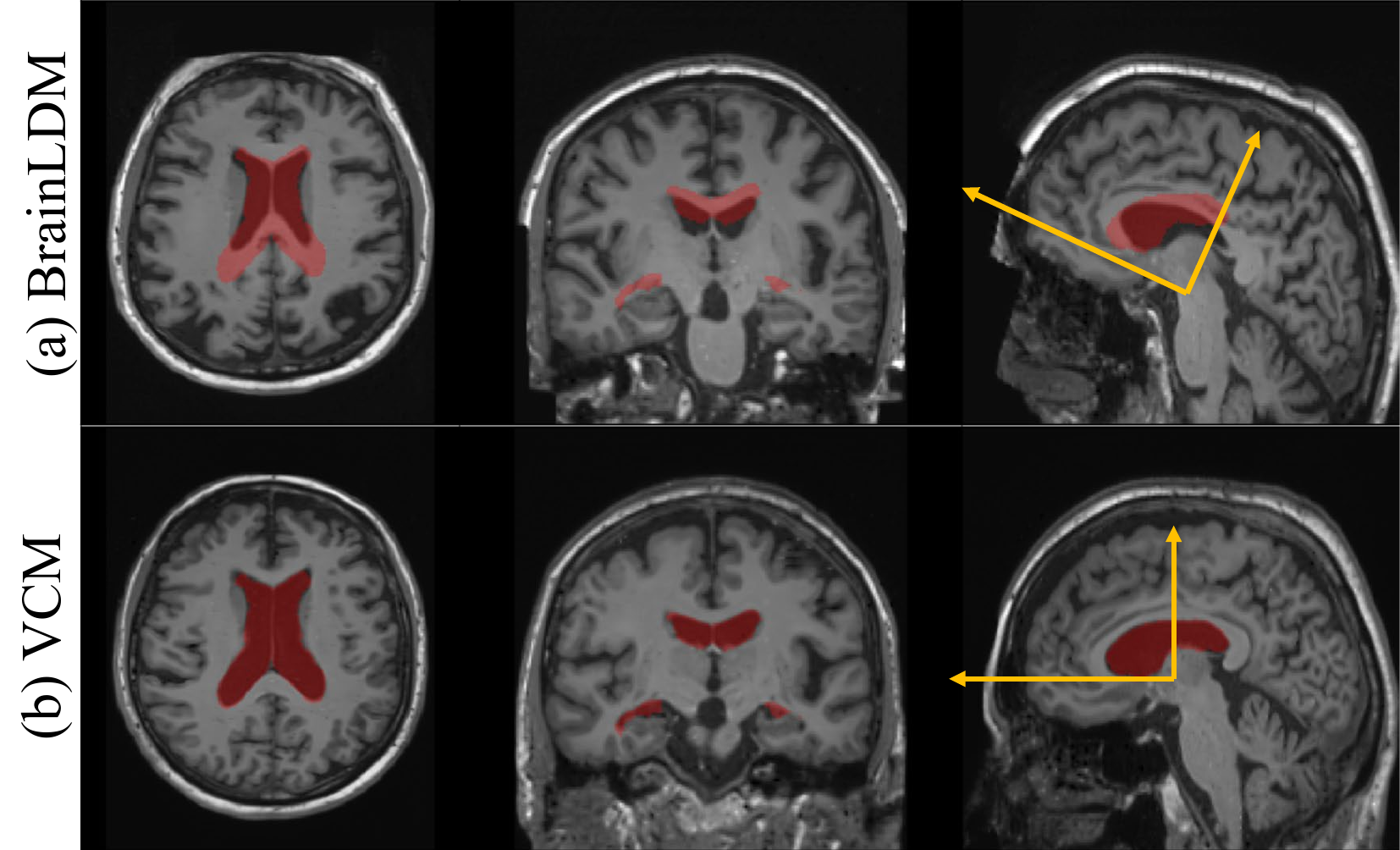}}
\caption{Spatially controlled output by VCM with the LV mask condition. (a) The original image synthesized by BrainLDM has a large displacement in terms of orientation with the given LV condition. (b) the guided output by VCM.}
\label{fig:registration-distance}
\end{figure}

\begin{figure}[t]
\centerline{\includegraphics[width=1.0\linewidth]{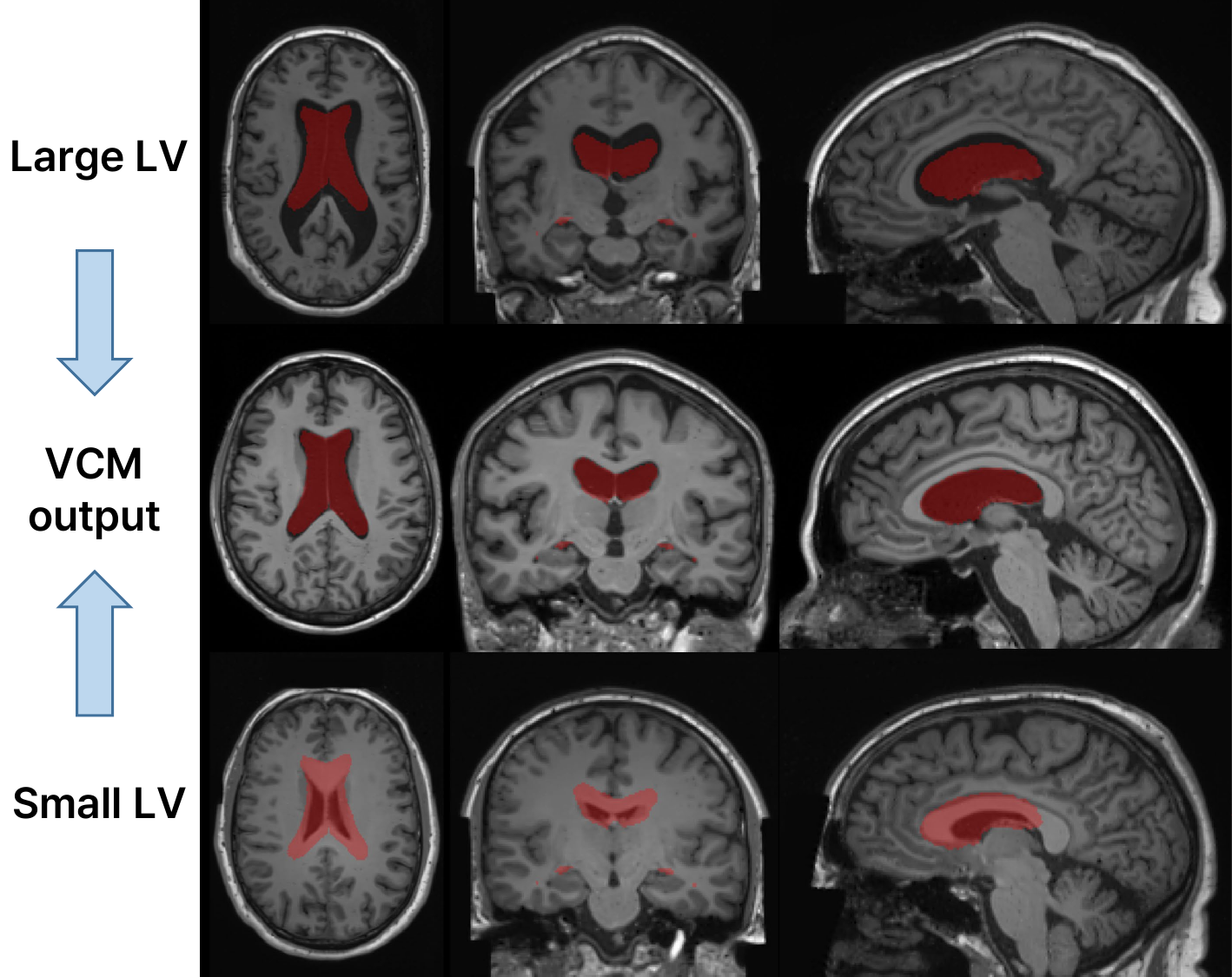}}
\caption{Examples of spatially controlled images by guiding outliers with VCM.}
\label{fig:Lv_extrem}
\end{figure}

We further analyze spatial controls through our VCM with lateral ventricle (LV) masks to generate T1w brain MRI scans. 
Through the following experiments and the corresponding analysis, we aim to understand what spatial control methods learn and act in guiding the generation process of pretrained diffusion models. 
To acquire an uncontrolled sample and a guided image with VCM, we sample both images copied from a single Gaussian noise.
Subsequently, only one noise is provided the spatial controls by VCM as we illustrate in \cref{fig:VCM_pipeline}. Finally, the predicted noises are decoded by the BrainLDM decoder to obtain T1w brain MRI scans. 

\begin{figure*}[!t]
\centerline{\includegraphics[width=1\linewidth]{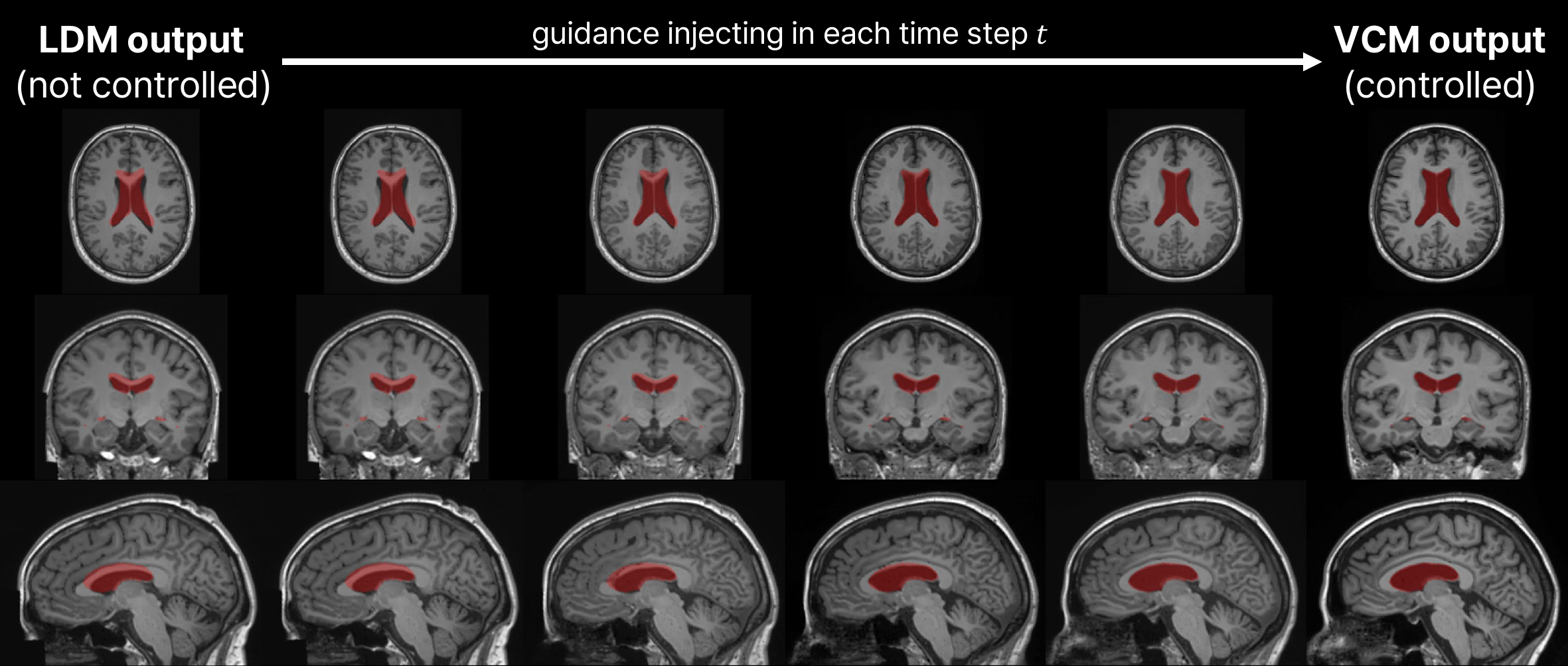}}
\caption{Visualization of spatial control guidance by VCM using the LV mask in each step of the diffusion process.}
\label{fig:Lv_timesteps}
\end{figure*}

\textbf{Appearance correction with given spatial conditions}. Synthetic medical images require anatomical fidelity for plausible image synthesis.
In this experiment, we assess whether VCM not only modifies the image based on given conditions, but also appropriately adjusts the overall appearance of the brain through orientation correction.
\cref{fig:registration-distance} shows the samples generated by BrainLDM and VCM at the top of BrainLDM, respectively.
If VCM learns spatial controls solely to create the LV in localized areas, the resulting output may appear unnatural, for instance, moving only the LV within the samples from BrainLDM.
However, our VCM properly adjusts the orientation of the entire brain corresponding to the given LV mask, as well as accurately guides the LV in the desired area. 
This correction of orientation indicates that spatial control methods leverage the capabilities of the pretrained models to generate the appropriate output within the brain image distribution.

\textbf{Spatial controls on outliers}. We investigate the influence and strength of spatial control methods by controlling outliers synthesized from BrainLDM. \cref{fig:Lv_extrem} shows the samples from BrainLDM using large and small scalar values for LV volume and the guided image by VCM with LV mask.
To make LV outlier images in the top and bottom rows of \cref{fig:Lv_extrem}, we input two extrapolating values of the conditioning variables for small and large LV volumes.
Although the LVs of the synthesized images from BrainLDM have significant differences with the given condition, the spatial control guides the generation process to match with the input mask, accurately.

Furthermore, we visualize the progress of spatial control of VCM by each time step in \cref{fig:Lv_timesteps}. 
In the early stage of generation, VCM mainly controls the generation steps to match the semantics of the LV and brain. 
After most of the semantics have been constructed by the diffusion process, the LV is not changed. Meanwhile, peripheral areas such as the cortical and subcortical areas are refined, appearing relevant structures with the given condition. 

\begin{figure}[]
\centerline{\includegraphics[width=1.0\linewidth]{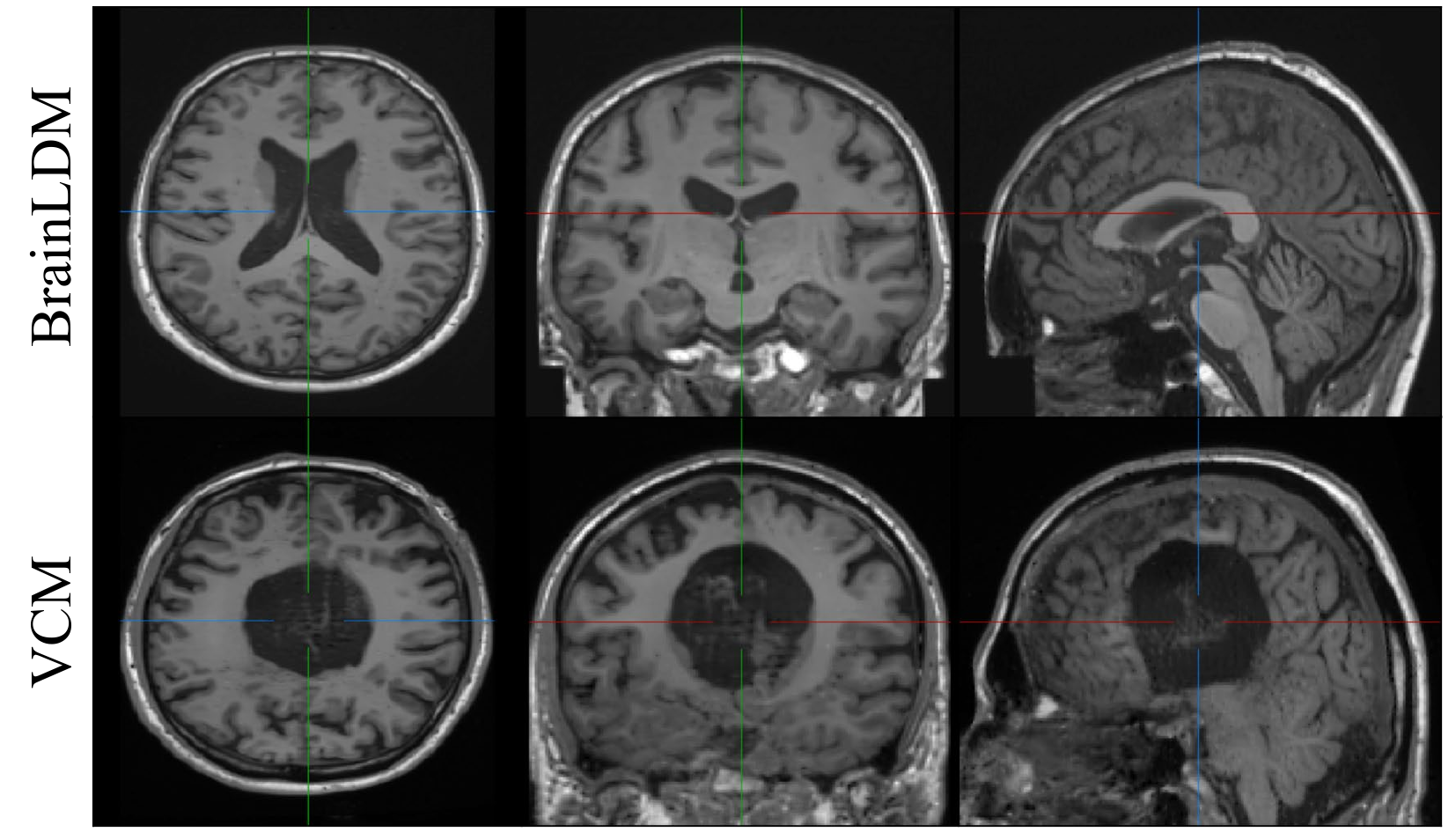}}
\caption{Synthetic image from an abnormal condition of the LV.}
\label{fig:adv_attack}
\end{figure}

\textbf{Extremely out-of-distribution conditions}. \label{subsec:adv}
Beyond various spatial control cases, we delve deeper into the investigation of robustness for VCM through abnormal input masks. 
To model the out-of-distribution condition, we replace the LV mask condition with a sphere-shaped mask. 
Since there is no training sample of the T1w brain MRI scan with the sphere-shaped LV, we can examine the robustness of our spatial control method.

\cref{fig:adv_attack} shows the generation results with VCM against an abnormal condition. 
Although the abnormal condition is unseen during VCM training, VCM controls the synthesized image to contain an LV structure localized with the given sphere mask. 
In addition, excluding the LV, the cortical area and skull shape show plausible appearances, as well as the background or the overall image intensity remains unaffected. 
However, the subcortical areas such as the corpus callosum and brain stem are destroyed. 
Unlike natural images whose image elements and compositions are highly diverse, medical images demand not only perceptually natural appearance but also anatomically correct structures. 
Consequently, the spatial control method fails to synthesize images under anatomically incorrect conditions. 
Thus, we suggest that researchers in the medical imaging field should be careful to use spatial control methods and properly evaluate the synthetic medical images.


\section{Experimental configuration details}
\label{supply:expr_details}

In this section, we provide a detailed description of the architecture and hyperparameters of VCM and the other methods for 3D implementations. In addition, we describe the details of the experiment setting.

\subsection{Datasets} The data used in our study are a total of 604 healthy T1w Brain MRI scans obtained from the Alzheimer’s Disease Neuroimaging Initiative (ADNI) database~\cite{jack2008alzheimer_ADNI}. 
All images were registered in MNI space using Advanced Normalization Tools \cite{avants2009advanced_ANTS}. 
For the 1D scalar input of BrainLDM, the volume values of LV and Brain were provided through SynthSeg outputs and demographic information such as sex and age was obtained from ADNI Image and Data Archive \cite{jack2008alzheimer_ADNI}. 
We normalize all T1w MRI scans and skull images to have a zero-to-one value range by min-max normalization from 0 to 255 values with intensity clipping.

\subsection{Implementation details}
\label{supply:impl}
\begin{figure}[]
\centerline{\includegraphics[width=1\columnwidth]{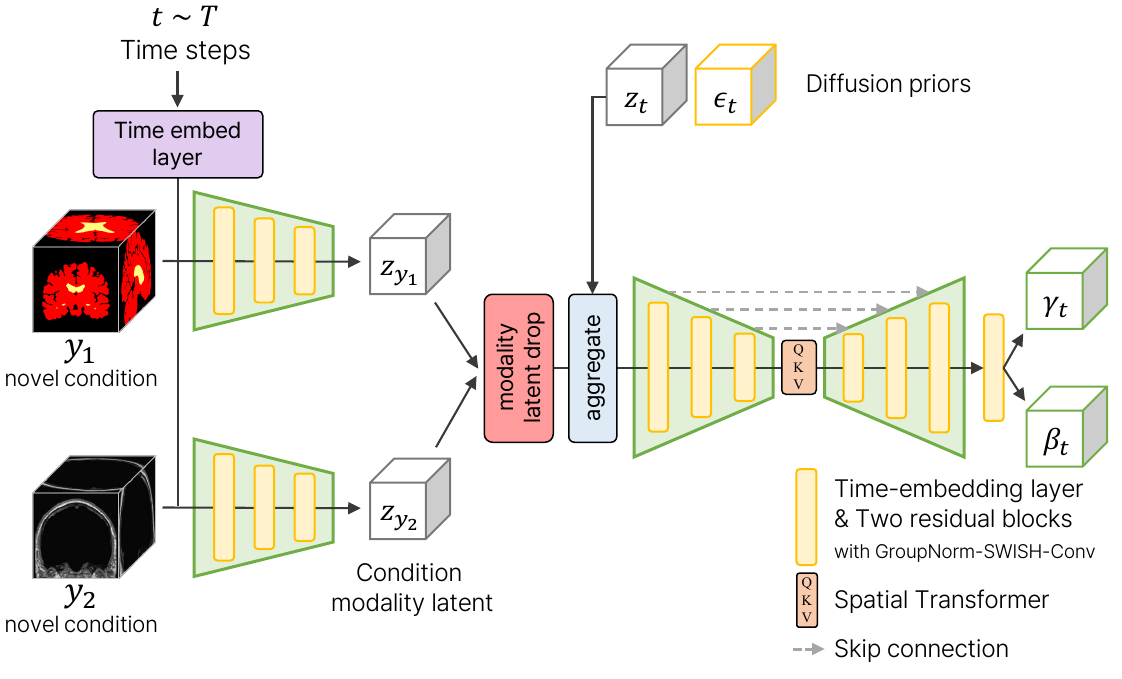}}
\caption{Details of VCM from \cref{fig:VCM_pipeline}. VCM takes diffusion priors, new conditions, and time steps as inputs, and outputs two modulation parameters $\{\gamma_t, \beta_t\}$ to modify the diffusion outputs.}
\label{fig:VCMdetails_modalitydrop}
\end{figure}

\textbf{Network Construction}. VCM consists of a time-conditioned asymmetric U-Net \cite{ronneberger2015unet} with a deeper encoder than the decoder and is implemented by modifying the \verb+DiffusionModelUNet+ MONAI generative \cite{pinaya2023generative_monaiGen}. 
We use 16 base channels for the encoder, with a channel multiplier of [1,2,3,4,8,16], whereas the decoder is used with the last three elements of the encoder multiplier, with the same base channels.
An MLP with two layers is used to embed each time step in the feature space (the purple box in \cref{fig:VCMdetails_modalitydrop}), while a single linear layer is utilized to project time embedding in the convolutional blocks in every VCM layer (the yellow box in \cref{fig:VCMdetails_modalitydrop}).
Two residual blocks are used at every level, composed of a time embedding layer, GroupNorm, SWISH activation functions, and a convolution layer.
Since the BrainLDM autoencoder maps a $1 \times 160 \times 224 \times 160$ T1w brain MRI scan to a latent vector with dimensions of $3 \times 20 \times 28 \times 20$, the last split layers of the VCM produce two modulation parameter tensors $\{\gamma_t, \beta_t\}$ in the same latent dimension.

\textbf{Hyperparameters}. For all experiments of \cref{sec:experiments}, the linear beta scheduler for noise scheduling is used in the beta range of 0.0015 and 0.0205. 
We train every method using DDPM \cite{ho2020denoising_ddpm} with 1000 diffusion steps, while DDIM \cite{song2022denoising_ddim} is used for inference with 200 diffusion steps. 
Our VCM runs over 10,000 epochs with a minibatch of 16 using AdamW optimizer \cite{loshchilov2017decoupled_adamw} and a base learning rate of $5\times10^{-5}$.
The $\lambda^{-1}$ of the loss for VCM in \cref{eq:total_loss} is $16 \times 3 \times 20 \times 28 \times 20$.
Axial flip augmentation is applied to improve generation performance according to Nichol, A. Q. and Dhariwal, P. \cite{nichol2021improvedddpm}, and linear learning rate annealing is employed to prevent unstable training in the early stage for all methods used in \cref{sec:experiments}. 

To analyze multimodal regularization techniques, we utilized two conditions: semantic map and partial image. During training VCM, the single semantic map condition, the single partial image condition, and the dual conditions appear with probability of 0.3, 0.3, and 0.4, respectively.
We use separate asymmetric parts of the encoder for each condition (see \cref{fig:VCMdetails_modalitydrop}) while keeping the number of parameters for VCM.

\begin{table}[!t]
\centering
\caption{Comparison of training time of conditional generation methods in 3D medical images.}
\label{tab:gpuTime}
\setlength{\tabcolsep}{3.3pt}
\fontsize{6.6}{7.5}\selectfont
\begin{tabular}{@{}c|cccc@{}}
\toprule
Method    & LDM (from scratch) & LDM (fine-tuning) & MCM            & T2I-Adapter \\ \midrule
\# params & 475.43M            & 553.19M           & 11.45M         & 447.26M     \\
GPU days  & 2.431              & 2.778             & 1.250          & 4.375       \\ \midrule
Method    & ControlNet         & ControlNet-LITE   & ControlNet-MLP & VCM         \\ \midrule
\# params & 193.88M            & 63.56M            & 9.03M          & 45.88M      \\
GPU days  & 4.375              & 3.472             & 3.264          & 3.882       \\ \bottomrule
\end{tabular}
\end{table}

\subsection{Comparison methods}
In the experiments of \cref{sec:1st_exp,sec:2nd_exp}, the \textit{input hint} CNN encoder and \verb+Pixelunshuffle+ are also implemented in volumetric spaces for the input downsampling of ControlNet \cite{zhang2023adding_controlNet} and T2I-Adapter \cite{mou2023t2iadapter}, respectively. 
In addition, we implement MCM-L \cite{ham2023modulating_mcm} which has the same complexity as our VCM but utilizes the MCM architecture. In the cases of ControlNet-LITE and ControlNet-MLP, we refer to their repository \cite{Yarish_2023_ctrlNetLITE} and replace all 2D learnable layers with 3D counterparts. For both variations of ControlNet, we employ the \textit{input hint} CNN encoder to maintain the downsampling methods of the original paper \cite{zhang2023adding_controlNet}. For fine-tuning BrainLDM \cite{pinaya2022brain_brainLDM} directly, we adjust the first convolution layer to adapt additional spatial conditions, while we directly input the conditions into the model for training an LDM from scratch, referring to the Dorjsembe, Zolnamar, et al. approach \cite{Dorjsembe_2024_medDDPM}.

In \cref{tab:gpuTime}, we provide GPU days of the comparison methods from \cref{sec:experiments}. Especially, the GPU days are analyzed by the same hyperparameters with 10,000 epochs with a minibatch of 16. For learning spatial controls within the 3D latent space, there is no significant difference among the competitive method ControlNet, lightweight variants of ControlNet (ControlNet-LITE and -MLP), and our VCM despite the differences in their model complexity.

\textbf{Super-resolution} For the super-resolution (SR) tasks in \cref{subsec:application,subsupply:SR}, we re-implement LIIF-3D by replacing the 2D learnable operation with the 3D counterparts of LIIF \cite{chen2021learning_LIIF}. 
Also, due to hardware limitations, we apply a patch-based approach to LIIF-3D to perform the SR in volumetric space. 
To learn continuous image representation as in the original research \cite{chen2021learning_LIIF}, the model was trained with random scales in $\times1-\times4$ and tested. 
In the case of InverseSR \cite{wang2023inversesr}, we utilize the decoder method and their official repository.

\section{Discussion for evaluation metrics}
In the experiments (\cref{sec:experiments}), we employ the FID \cite{heusel2018gans_FID} and LPIPS \cite{zhang2018unreasonable_LPIPS} to measure image quality and diversity, respectively, since they are widely used in both 2D synthetic natural images \cite{ham2023modulating_mcm} and 3D medical images \cite{Dorjsembe_2024_medDDPM}.  
However, we discuss that these metrics are insufficient for analyzing synthetic medical images.

\begin{table}[]
\centering
\setlength{\tabcolsep}{6pt}
\fontsize{7.5}{8.5}\selectfont
\caption{Quantitative evaluation of the synthetic images on the validation dataset and the results of VCM with abnormal LV mask.}
\label{tab:eval-err}
\begin{tabular}{@{}c|c|ccc@{}}
\toprule
Methods            & \# train data & FID 2D↓ & FID 3D↓ & LPIPS↑ \\ \midrule
BrainLDM           & 31,740 \cite{Sudlow2015-fz__UKbioBank}        & 84.511  & 2.247   & 0.340  \\
LDM (fine-tuning)  & 50 \cite{jack2008alzheimer_ADNI}           & 83.619  & 2.497   & 0.314  \\
LDM (from scratch) & 50 \cite{jack2008alzheimer_ADNI}            & 92.404  & 4.845   & 0.476  \\
VCM (sphere masks) & -             & 81.444  & 7.515   & 0.331  \\ \bottomrule
\end{tabular}
\end{table}

\begin{figure}[]
\centerline{\includegraphics[width=1.0\linewidth]{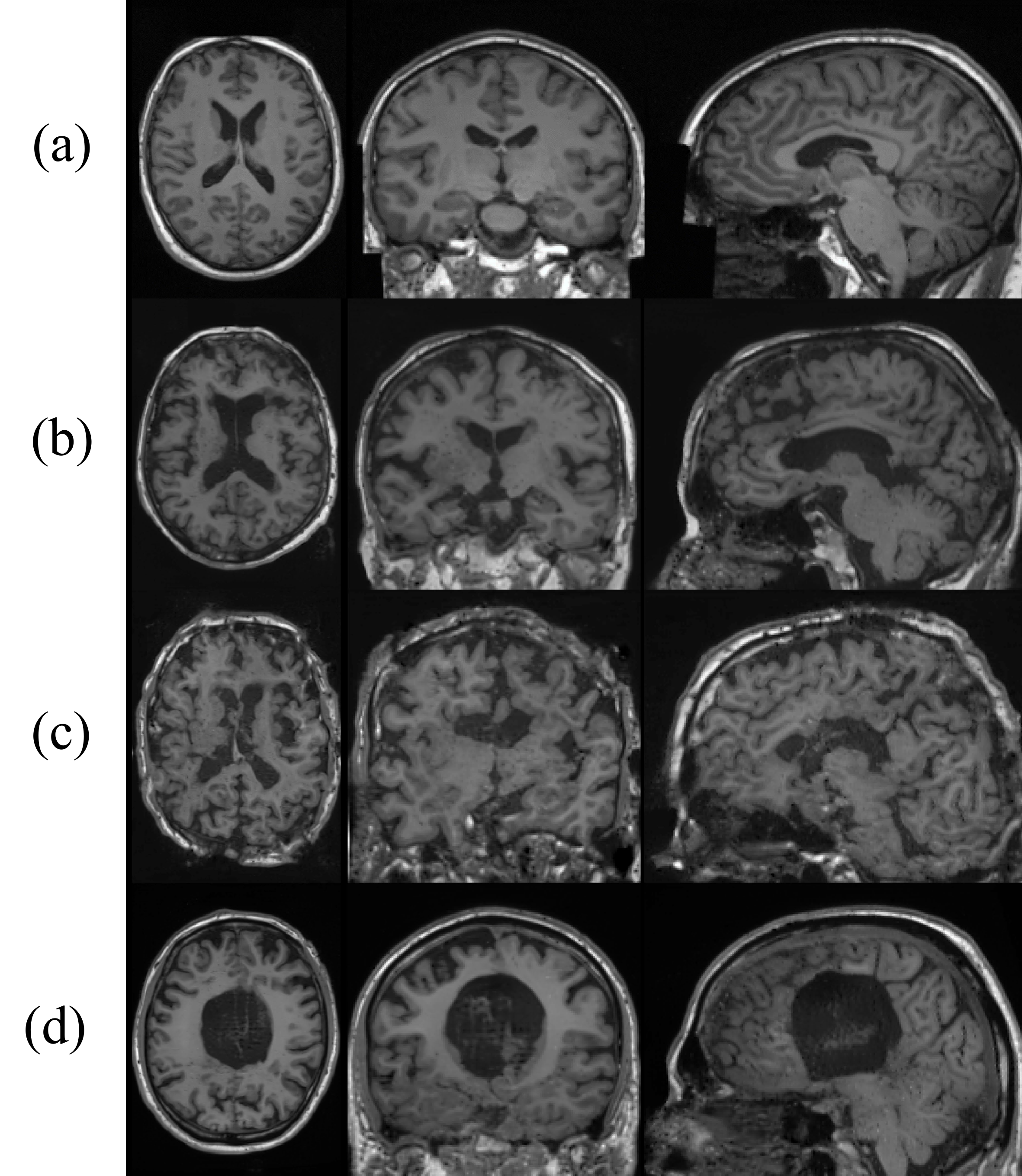}}
\caption{Examples of the generated output in \cref{tab:eval-err}. (a) the synthesized output of BrainLDM. (b) and (c) shows the samples from LDM trained by fine-tuning and from scratch, respectively. (d) shows the result of the abnormal condition of \cref{subsec:adv}.}
\label{fig:eval-err-vis}
\end{figure}

As illustrated in \cref{tab:eval-err}, samples from BrainLDM have $84.511$ 2D FID from real scans in the validation dataset of \cref{sec:experiments}. 
However, even the wrongly synthesized images with noisy white matter in \cref{fig:eval-err-vis} (b) have a smaller 2D FID of $83.619$, which means closer to the real image distribution. 
In addition, although 3D FID is larger than BrainLDM, it has a subtle difference, which could be misobserved in plausible synthetic images.
Moreover, for the synthetic image conditioned with sphere LV masks (\cref{fig:eval-err-vis} (d)), the 2D FID is even smaller than the BrainLDM, while the 3D FID is large enough to judge that the synthetic image is far from the real image distribution.
The evaluation results of the severely destroyed synthetic images (\cref{fig:eval-err-vis} (c)), which have a much more degraded and noisy appearance, finally show distinguishable large values in both 2D and 3D FID.
In the case of LPIPS metrics, large values, which means high diversity in natural images, are insufficient to explain the diversity of generative models for medical images, since a severely abnormal appearance brings large LPIPS values, as shown in \cref{fig:eval-err-vis}.

Based on the quantitative evaluation in \cref{tab:eval-err}, the following discussion points arise.
First, qualitative evaluation is important in assessing the validity of synthetic medical images.
Additionally, to evaluate the quality of 3D medical images, 2D FID which is widely used to measure the distance with 2D slices is less effective than 3D FID which refers to the entire 3D image.
Finally, when large FID and LPIPS are measured, researchers should keep in mind that this means that the model synthesizes destroyed medical images beyond anatomically unnatural images or diverse images.
Above all, innovative and effective evaluation metrics for medical images are essential for advancing research on medical image generative models.

\section{Acknowledgment}
This work was supported by Institute for Information \& communications Technology Promotion(IITP) grant funded by the Korea government(MSIT) (No.00223446, Development of object-oriented synthetic data generation and evaluation methods).

Data collection and sharing for this project was funded by the Alzheimer's Disease Neuroimaging Initiative (ADNI) (National Institutes of Health Grant U01 AG024904) and DOD ADNI (Department of Defense award number W81XWH-12-2-0012). ADNI is funded by the National Institute on Aging, the National Institute of Biomedical Imaging and Bioengineering, and through generous contributions from the following: AbbVie, Alzheimer’s Association; Alzheimer’s Drug Discovery Foundation; Araclon Biotech; BioClinica, Inc.; Biogen; Bristol-Myers Squibb Company; CereSpir, Inc.; Cogstate; Eisai Inc.; Elan Pharmaceuticals, Inc.; Eli Lilly and Company; EuroImmun; F. Hoffmann-La Roche Ltd and its affiliated company Genentech, Inc.; Fujirebio; GE Healthcare; IXICO Ltd.; Janssen Alzheimer Immunotherapy Research \& Development, LLC.; Johnson \& Johnson Pharmaceutical Research \& Development LLC.; Lumosity; Lundbeck; Merck \& Co., Inc.; Meso Scale Diagnostics, LLC.;  NeuroRx Research; Neurotrack Technologies; Novartis Pharmaceuticals Corporation; Pfizer Inc.; Piramal Imaging; Servier; Takeda Pharmaceutical Company; and Transition Therapeutics. The Canadian Institutes of Health Research is providing funds to support ADNI clinical sites in Canada. Private sector contributions are facilitated by the Foundation for the National Institutes of Health (www.fnih.org). The grantee organization is the Northern California Institute for Research and Education, and the study is coordinated by the Alzheimer’s Therapeutic Research Institute at the University of Southern California. ADNI data are disseminated by the Laboratory for Neuro Imaging at the University of Southern California. 

\end{document}